\documentclass{article}

\PassOptionsToPackage{numbers,compress}{natbib}

\usepackage[final,sglblindworkshop]{neurips_2025}
\usepackage[utf8]{inputenc}
\usepackage[T1]{fontenc}
\usepackage{hyperref}
\usepackage{url}
\usepackage{booktabs}
\usepackage{amsfonts}
\usepackage{amsmath}
\usepackage{nicefrac}
\usepackage{microtype}
\usepackage{xcolor}
\usepackage{graphicx}
\usepackage{subcaption}
\usepackage{wrapfig}
\usepackage{multirow}
\usepackage{amsthm}
\usepackage{colortbl}
\usepackage{float}

\newcommand{\SL}[1]{\textcolor{black}{#1}}

\renewcommand{\acksection}{\section*{Acknowledgments}}

\makeatletter
\renewcommand{\And}{
  \end{tabular}\hfil\linebreak[0]\hfil
  \begin{tabular}[t]{c}\bf\rule{\z@}{18\p@}\ignorespaces
}
\makeatother

\title{EnfoPath: Energy-Informed Analysis of \\ Generative Trajectories in Flow Matching}

\workshoptitle{EurIPS 2025 Workshop on Principles of Generative Modeling (PriGM)}

\author{
  Ziyun Li\thanks{Corresponding author} \\
  KTH Royal Institute of Technology\\
  Stockholm, Sweden \\
  \texttt{ziyli@kth.se; liziyun2014@gmail.com} \\
  \And
  Ben Dai \\
  Chinese University of hong kong\\
  Hong Kong, China \\
  \texttt{ben.dai@cuhk.edu.hk} \\
  \AND
  Huancheng Hu \\
  Hasso Plattner Institute\\
  Potsdam, Germany \\
  \texttt{huancheng.hu@hpi.de} \\
  \And
  Henrik Boström \\
  KTH Royal Institute of Technology\\
  Stockholm, Sweden \\
  \texttt{bostromh@kth.se} \\
  \And
  Soon Hoe Lim \\
  KTH Royal Institute of Technology \& Nordita\\
  Stockholm, Sweden \\
  \texttt{shlim@kth.se} \\
}

\begin{document}

\maketitle

\begin{abstract}
Flow-based generative models synthesize data by integrating a learned velocity field from a reference distribution to the target data distribution. Prior work has focused on endpoint metrics (e.g., fidelity, likelihood, perceptual quality) while overlooking a deeper question: what do the sampling trajectories reveal?
Motivated by classical mechanics, we introduce kinetic path energy (KPE), a simple yet powerful diagnostic that quantifies the total kinetic effort along each generation path of ODE-based samplers.  Through comprehensive experiments on CIFAR-10 and ImageNet-256, we uncover two key phenomena: (\emph{i}) higher KPE predicts stronger semantic quality, indicating that semantically richer samples require greater kinetic effort, and (\emph{ii}) higher KPE inversely correlates with data density, with informative samples residing in sparse, low-density regions. Together, these findings reveal that semantically informative samples naturally reside on the sparse frontier of the data distribution, demanding greater generative effort. Our results suggest that trajectory-level analysis offers a physics-inspired and interpretable framework for understanding generation difficulty and sample characteristics.

\noindent\textbf{Code:} \url{https://github.com/ziyunli-2023/EnfoPath-FM}.

\end{abstract}

\section{Introduction}
\label{sec:introduction}
Flow-based generative models~\cite{liu2022flow,lipman2022flow,tong2024improving,albergo2023stochastic,chen2018neural} synthesize high-fidelity images by continuously transforming noise into data. Despite extensive research on architectures~\cite{peebles2023scalable,ma2024sit,rombach2022high}, training objectives~\cite{ho2022classifier,pooladian2023multisample}, and sampling efficiency~\cite{song2020denoising,lu2022dpm,karras2022elucidating}, existing work focuses almost exclusively on endpoint metrics~\cite{heusel2017gans,parmar2022aliased}, such as FID, likelihood, perceptual quality, while ignoring the sampling trajectory itself. 
Only a few works consider trajectory behavior: AMED-Solver \cite{zhou2024fast} exploits trajectory smoothness to allow very large integration steps, PFDiff \cite{wang2025pfdiff} uses past and future score information to approximate local trajectory curvature, and FlowGuard \cite{li2025flowguard} monitors velocity and curvature in flow-matching trajectories to reject unreliable paths.
Yet in  \SL{deterministic ODE-based samplers~\cite{song2020denoising,karras2022elucidating} of flow matching, each generated sample is obtained as the result of a realization of a  path \( \{x(t)\}_{t \in [0,1]} \) governed by \( \dot{x}(t) = v_\theta(x(t), t) \), where \(v_\theta\) is the learnt velocity field and $x(0) \sim p_0$ for some reference (noise) distribution $p_0$ (typically  standard Gaussian).} This raises a fundamental question: \emph{Can the trajectory itself reveal sample quality, generation difficulty, and data characteristics?}

\SL{To explore this, we adopt a trajectory-wise perspective and draw inspiration from classical mechanics \cite{goldstein2002classical,feynman1965quantum} and recent path integral approaches to generative modeling~\cite{hirono2024pathintegral,zhang2023path}.}
\SL{The central quantity of interest is}  the \SL{\textit{kinetic path energy (KPE)}:}
\[
E = \tfrac{1}{2} \int_0^1 \|v_\theta(x(t), t)\|^2\, dt,
\]
\SL{which we also refer to as simply {\it trajectory energy}. Viewing each sample as a free particle traversing from noise to data, we can interpret $E$ as a measure of the total kinetic cost of this journey.} Though not physically grounded in the strict sense, this lens reveals striking patterns. Through comprehensive experiments on CIFAR-10~\cite{krizhevsky2009learning} and ImageNet-256~\cite{deng2009imagenet}, we have two key findings:
\textit{(1) Higher energy, stronger semantics.} \SL{Higher-$E$ trajectories} consistently produce semantically richer, more class-specific images (CLIP score/margin~\cite{radford2021learning}, \autoref{sec:energy_semantic}).
In physics, reaching a precise target requires sustained, directed force, \SL{incurring a high} kinetic cost. Analogously, generating semantically rich, class-specific images demands that the model maintain \SL{high velocity magnitudes along the trajectory, accumulating more energy.}
\textit{(2) Higher energy, lower density.} High-$E$ samples fall in sparse, low-density regions of the data manifold (\autoref{sec:energy_density}). 
A particle with higher \SL{kinetic energy travels farther from its origin.}
Analogously, reaching atypical, \SL{lower-density} regions, distant from the training distribution's center, requires the model to have higher velocities throughout the trajectory, thus higher energy.
These findings establish \SL{KPE} as a dual indicator: it predicts both semantic strength and sample rarity, revealing that high-quality, informative samples naturally inhabit low-density regions.

We summarize our contributions as follows:
\begin{itemize}
  \item We introduce \SL{KPE}  as a simple yet powerful diagnostic that quantifies the kinetic cost of \SL{ODE-based generation process in flow matching} through a physics-inspired lens.
 \item \SL{Our empirical results show that: (i) higher KPE predicts higher semantic quality: class-specific samples demand greater kinetic cost to generate, (ii) higher KPE indicates lower data density: rare samples occupy regions \SL{that require} greater kinetic cost to reach.}
\end{itemize}

\section{Kinetic Analogy and Trajectory Energy}
\label{sec:theory}
\subsection{Physical Motivation}

\SL{In classical mechanics, the evolution of a system is characterized by the action functional}~\citep{goldstein2002classical,feynman1965quantum}
\[
S[x(\cdot)] = \int_{t_0}^{t_1} L(x(t), \dot{x}(t), t) \, dt,
\]
\SL{defined over the time interval \([t_0, t_1] \), where \(L(x(t), \dot{x}(t), t) =  T(\dot x(t)) - V(x(t))   \) is the Lagrangian, given by the difference between the kinetic energy \(T(\dot x)\)  and the potential energy \(V(x)\).} This formulation embodies Hamilton's principle of least action~\citep{goldstein2002classical} and Feynman's path integral framework~\citep{feynman1965quantum}.
For a free particle (i.e., \SL{when  \(V(x) = 0\)}), the \SL{action functional reduces to the kinetic term}:
\[
S_{\mathrm{free}} = \int_{t_0}^{t_1} \tfrac{1}{2}\|\dot x(t)\|^2\, dt.
\]
This kinetic form is a fundamental example of an action functional in physics. Inspired by similar analogies in generative modeling~\citep{hirono2024pathintegral,zhang2023path}, we adopt a kinetic framework to define trajectory-level diagnostics to \SL{gain insight into the generation process in flow matching.}

\subsection{Kinetic Cost Definition}
We interpret the flow matching sampling process as a particle moving through a velocity field. The sampling trajectory is governed by the learned velocity field \(v_\theta(x, t)\) via the ODE~\cite{liu2022flow,lipman2022flow,song2020denoising,chen2018neural}:
\[
\frac{dx}{dt} = v_\theta(x(t), t), \quad t \in [0, 1], \quad \SL{x(0) \sim \mathcal{N}(0, I),}
\]
which describes the probability flow~\cite{song2020score} from noise to data distribution. The trajectory represents the path of a particle driven by the velocity field.
Inspired by classical mechanics~\cite{goldstein2002classical}, we define \textit{kinetic path energy} \(E\) as:
\[
E := \tfrac{1}{2} \int_0^1 \|v_\theta(x(t), t)\|^2\, dt,
\]
where we adopt the convention of \textit{unit mass (\(m=1\))}, \SL{standard in} the free particle action formulation. 

\textbf{Physical interpretation.}
The quantity \(E\) encapsulates the cumulative kinetic \SL{cost incurred during the sampling process.} Conceptually, it quantifies the ``energy'' required to transport a sample from the noise distribution to the data manifold. A higher \(E\) signifies that the model employs greater velocity magnitudes on average, reflecting a more energetically demanding generation process.

\textbf{Important clarification.}
We emphasize that \(E\) is not \SL{strictly} a physical energy, but rather a \textit{kinetic-inspired metric} designed to quantify trajectory complexity. This metric is straightforward to compute: it only requires recording the velocity field magnitudes during standard ODE integration, incurring negligible computational overhead. \SL{Moreover, the expected KPE matches the Benamou–Brenier dynamic formulation~\citep{benamou2000computational} of the 2-Wasserstein transport cost when the learned flow realizes the optimal transport~\citep{tong2024improving,pooladian2023multisample}, grounding our diagnostic in optimal transport theory~\citep{villani2009optimal} rather than mere physical analogy.}

\section{Experiments}
\label{sec:experiments}
\subsection{Experimental Setup}
\SL{To explore the properties of the sampling trajectory, we} evaluate on CIFAR-10 \cite{krizhevsky2009learning} using a pretrained OTCFM model \cite{tong2024improving}, and on ImageNet-256 \cite{deng2009imagenet} using two pretrained models: DiT-B \cite{peebles2023scalable} (shortcut~\cite{peebles2023scalable}) and SiT-XL/2 \cite{ma2024sit} with Stable Diffusion VAE for latent encoding.

\subsection{\SL{KPE ${E}$ vs. Semantic Strength}}
\label{sec:energy_semantic}
\textbf{Setup and Metrics.} 
We generate 5K ImageNet-256 samples using \SL{the forward Euler sampler} with \SL{Classifier-Free Guidance (CFG)} \cite{ho2022classifier} scales of 1.0, 1.5, and 4.0. \SL{CFG is a common technique in diffusion and flow-based models that interpolates between unconditional and class-conditional generation to control the strength of semantic conditioning—larger scales typically yield sharper, more class-consistent images at the cost of diversity. While our initial setup considers trajectories from the unconditional vector field $v_\theta(x,t)$, we also examine CFG sampling, where a constructed mixture field amplifies conditional guidance -- allowing us to study how stronger semantic alignment influences  KPE and sample quality.} Trajectories are grouped into low (0--33\%), mid (33--67\%), and high (67--100\%) energy bins, and evaluated using ${E}$, CLIP score, and CLIP margin~\cite{radford2021learning}.
\textbf{CLIP score} quantifies semantic alignment as the maximum cosine similarity (scaled by 100) between normalized image and text features for the true class (e.g., ``a photo of a golden retriever'').  
\textbf{CLIP margin} measures semantic discriminability as the difference between the similarity to the true class and the highest similarity to competing classes:
$
\text{Margin} = \text{Sim}_{\text{true}} - \max_{c \in \mathcal{C}_{\text{others}}} \text{Sim}(c),
$
where $\mathcal{C}_{\text{others}}$ represents competing classes. Higher margins indicate stronger class-specific semantics.

\begin{figure}[h]
  \centering
  \includegraphics[width=0.93\linewidth]{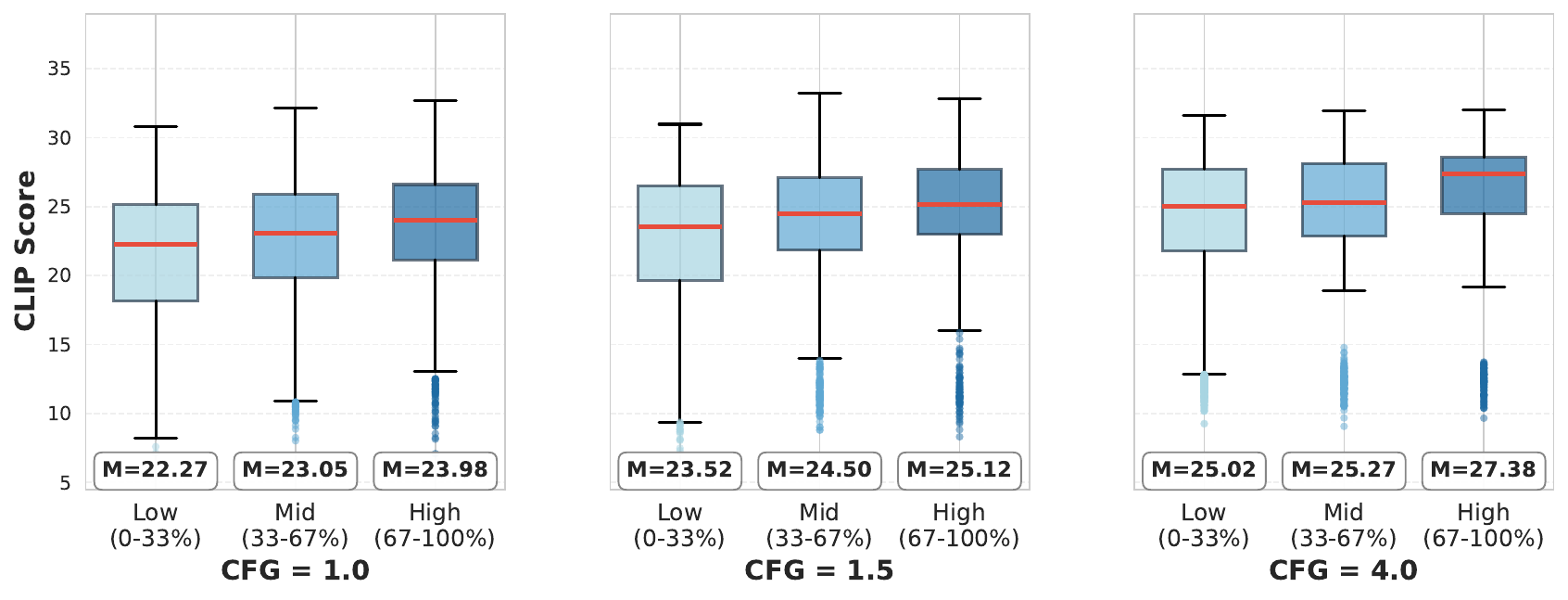}
  \caption{Trajectory energy ($E$) positively correlates with CLIP score across CFG settings.}
  \label{fig:energy_clip_scatter}
\end{figure}

\begin{figure}[h]
  \centering
  \includegraphics[width=0.93\linewidth]{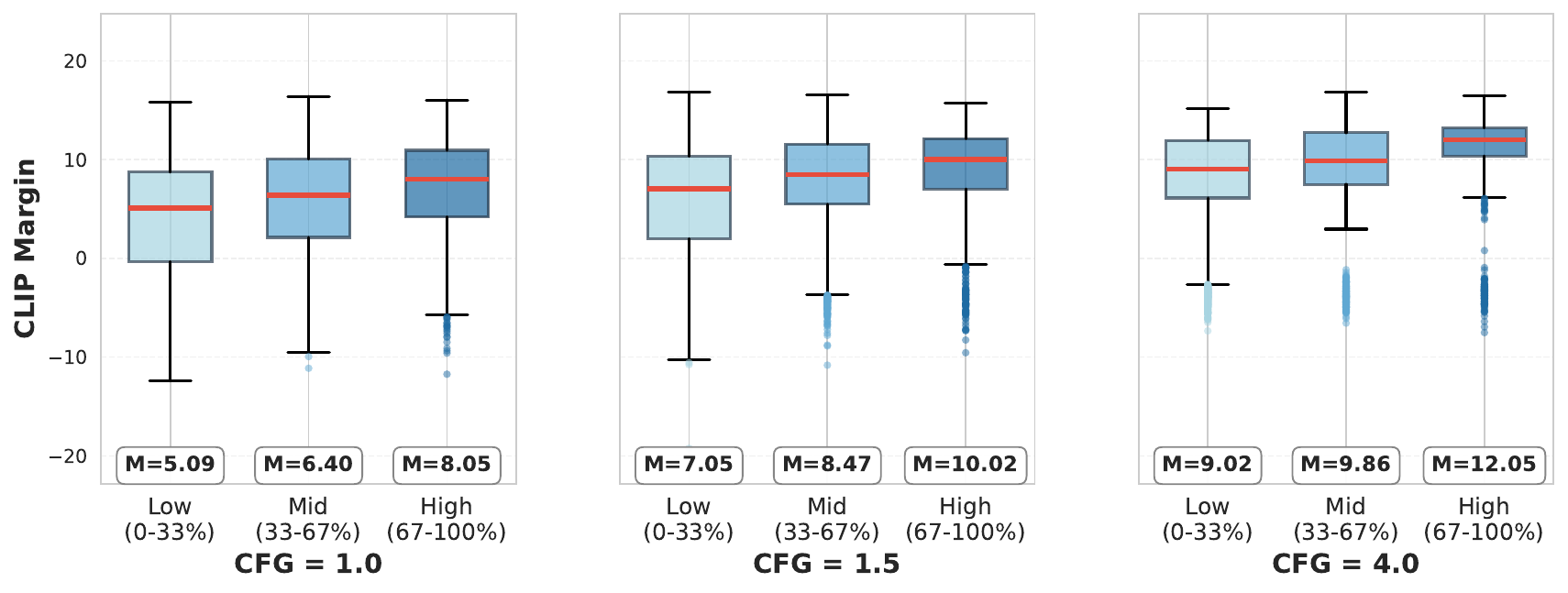}
  \caption{Trajectory energy ($E$) positively correlates with CLIP margin across CFG settings.}
  \label{fig:energy_clip_margin_scatter}
\end{figure}

\textbf{Results.} 
\autoref{fig:energy_clip_scatter} and \autoref{fig:energy_clip_margin_scatter} \SL{show  that both CLIP score and CLIP margin increase with ${E}$ across different CFG settings.}
For instance, at CFG=1.5, the median CLIP score increases from 23.52 (low-energy) to 25.12 (high-energy), and the median CLIP margin rises from 7.05 to 10.02.
Additionally, \autoref{tab:clip_score} and \autoref{tab:clip_margin} compare low-energy and high-energy groups.  Independent two-sample $t$-tests ($p < 0.01$) confirm statistical significance, with large effect sizes (Cohen's $d$) emphasizing practical relevance. These findings demonstrate that higher-energy trajectories produce images with clearer and more accurate semantics.
\SL{More visualization examples are provided in the Appendix.}

\textbf{Conclusion.} 
\emph{Higher  $E$ consistently correlates with stronger semantic alignment and discriminability.}

\begin{table}[H]
  \centering
  \caption{CLIP Score: Energy Group Comparison across CFG Scales (High vs. Low)}
  \label{tab:clip_score}
  \begin{tabular}{@{}ccccccc@{}}
  \toprule
  \textbf{CFG} & \textbf{Low Energy} & \textbf{High Energy} & \textbf{$\Delta\mu$} & \textbf{$t$-statistic} & \textbf{$p$-value} & \textbf{Cohen's $d$} \\
  \textbf{Scale} & $\mu \pm \sigma$ & $\mu \pm \sigma$ & & & & \\
  \midrule
  1.0 & $21.22 \pm 5.35$ & $23.43 \pm 4.44$ & $+2.21$ & $11.55$ & $3.75 \times 10^{-30}$ & $0.450$ \\
  1.5 & $21.87 \pm 5.99$ & $24.62 \pm 4.29$ & $+2.75$ & $13.54$ & $2.02 \times 10^{-40}$ & $0.527$ \\
  4.0 & $23.23 \pm 5.89$ & $25.87 \pm 4.39$ & $+2.64$ & $13.06$ & $7.99 \times 10^{-38}$ & $0.509$ \\
  \bottomrule
  \end{tabular}
  
  \footnotesize
  \textit{Notes:} All comparisons are statistically significant at $p < 0.001$. Cohen's $d$ values indicate medium to large effect sizes. Low energy: 0--33\% percentile; High energy: 67--100\% percentile. $n = 1320$ per group.
  \end{table}

  \begin{table}[H]
  \centering
  \caption{CLIP Margin: Energy Group Comparison across CFG Scales (High vs. Low)}
  \label{tab:clip_margin}
  \begin{tabular}{@{}ccccccc@{}}
  \toprule
  \textbf{CFG} & \textbf{Low Energy} & \textbf{High Energy} & \textbf{$\Delta\mu$} & \textbf{$t$-statistic} & \textbf{$p$-value} & \textbf{Cohen's $d$} \\
  \textbf{Scale} & $\mu \pm \sigma$ & $\mu \pm \sigma$ & & & & \\
  \midrule
  1.0 & $4.15 \pm 5.99$ & $7.09 \pm 5.00$ & $+2.94$ & $13.70$ & $2.42 \times 10^{-41}$ & $0.534$ \\
  1.5 & $5.66 \pm 6.17$ & $8.93 \pm 4.54$ & $+3.27$ & $15.48$ & $8.40 \times 10^{-52}$ & $0.603$ \\
  4.0 & $7.44 \pm 5.95$ & $10.82 \pm 4.40$ & $+3.38$ & $16.59$ & $7.21 \times 10^{-59}$ & $0.646$ \\
  \bottomrule
  \end{tabular}
  
  \footnotesize
  \textit{Notes:} All comparisons are statistically significant at $p < 0.001$. Cohen's $d$ values indicate medium to large effect sizes. Low energy: 0--33\% percentile; High energy: 67--100\% percentile. $n = 1320$ per group.
  \end{table}

\subsection{KPE ${E}$ vs. Data Density}
\label{sec:energy_density}

\textbf{Setup and Metrics.}
We analyze the correlation \SL{between $E$ and estimated sample densities using the} OTCFM \cite{tong2024improving} (on CIFAR-10) and Shortcut \cite{frans2025onestep} (on ImageNet256). We generate 2,000 samples via Euler solver ($N=10, 50, 150$ steps) and extract \textit{low-dimensional} image features, including channel statistics, texture, and edge descriptors, resulting in 22 features per image. PCA reduces these to 2D embeddings (explained variance $>85\%$). Density is estimated using $k$-NN ($k=50$) and Gaussian KDE with Scott's bandwidth.
We use Spearman's $\rho$ to quantify monotonic correlation, and Cliff's $\delta$ to measure effect size between high-energy (top 20\%) and low-energy (bottom 20\%) groups, validated using the Mann-Whitney $U$ test on CIFAR10 data.

\begin{figure}[h]
  \vspace{-5.2mm}
  \centering
  \begin{subfigure}[t]{0.69\linewidth}
    \centering
    \includegraphics[width=\linewidth,clip,trim=0cm 0.2cm 0cm 1.0cm]{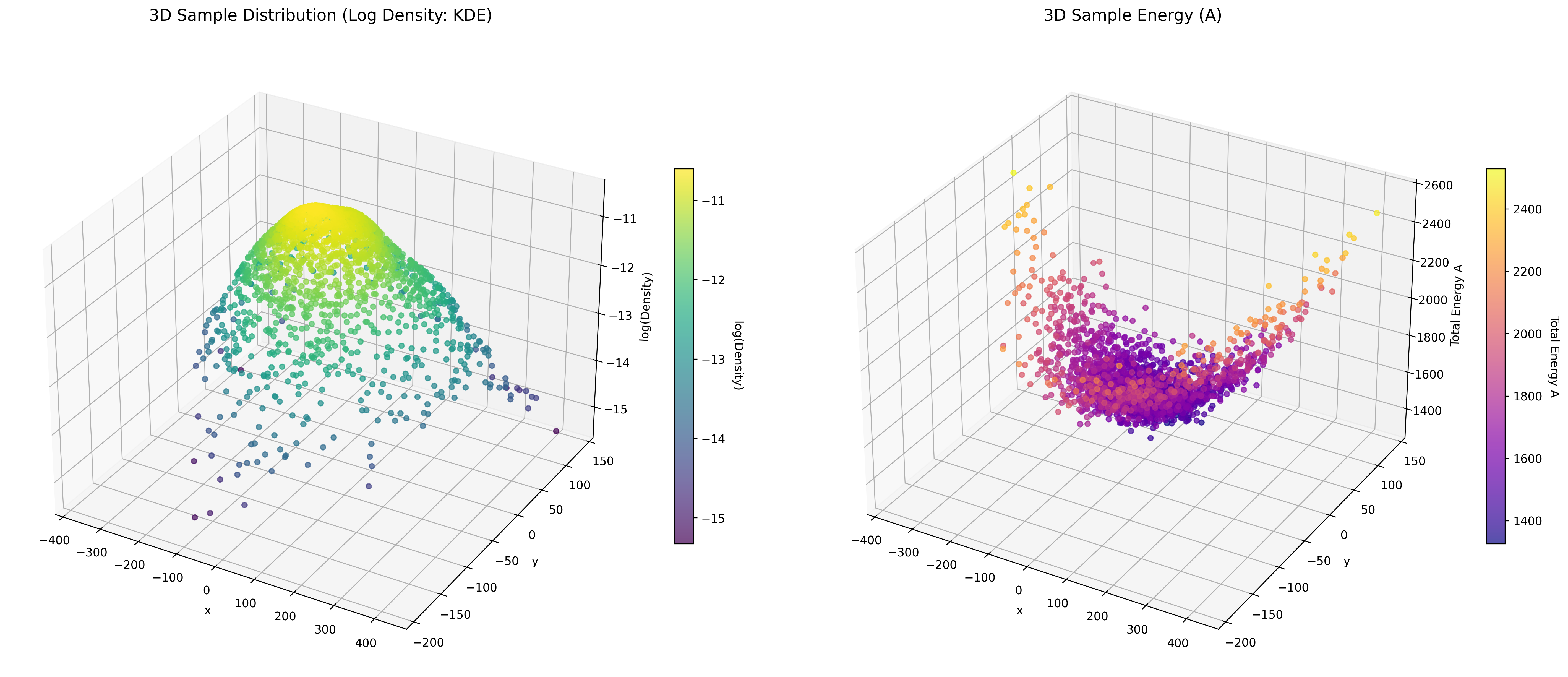}
    \caption{KPE vs. density KDE surface}
    \label{fig:a_density_3d}
  \end{subfigure}\hfill
  \begin{subfigure}[t]{0.31\linewidth}
    \centering
    \includegraphics[width=\linewidth,clip,trim=0cm 0.2cm 0cm 1.0cm]{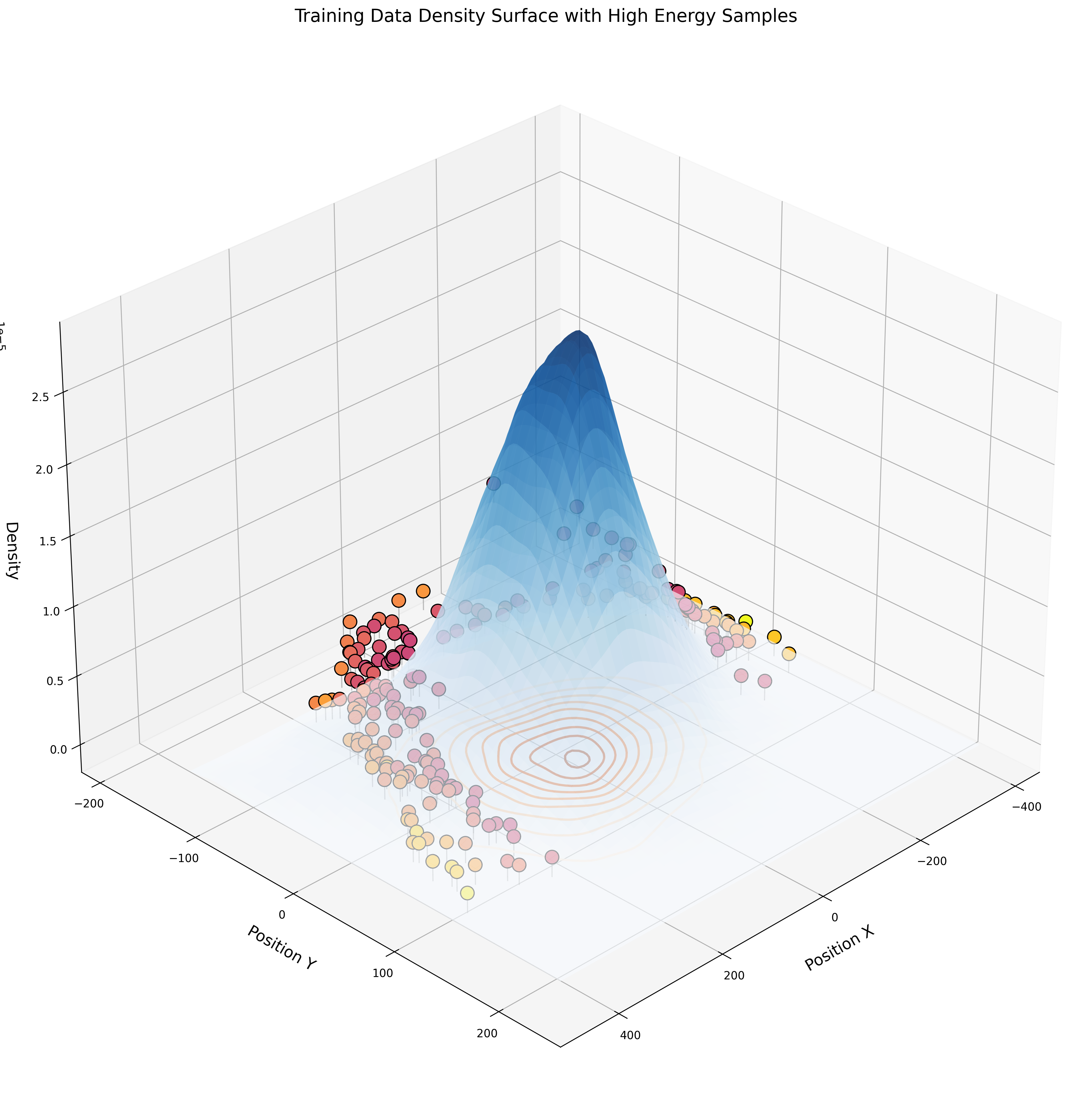}
    \caption{High-energy region highlights}
    \label{fig:high_energy_region}
  \end{subfigure}
  \caption{(a) 3D surfaces of $\log(\text{density})$ (left) and trajectory energy $E$ (right) (CIFAR-10, 150 steps) show high-density regions align with low energy. 
  (b) Top 10\% highest-energy samples consistently fall in low-density regions, confirming the inverse energy-density correlation.}
  \label{fig:a_density_pair}
\end{figure}

\textbf{Qualitative Results.} We observe \textit{a negative correlation between  ${E}$ and data density}. \autoref{fig:a_density_3d} shows 3D visualizations of $\log(\text{density})$ (left) and  ${E}$ (right). 
The surfaces exhibit opposite shapes: high-density areas align with low energy, and low-density areas align with high energy.
In \autoref{fig:high_energy_region}, the top 10\% highest-energy generated samples are projected onto the training density surface, consistently falling in low-density regions. \SL{This  empirically illustrates} the inverse energy–density relationship.

\begin{figure}[h]
  \vspace{-6mm}
  \centering
  \tiny
  \noindent
  \begin{minipage}[c]{0.63\linewidth}
    \centering
    \setlength{\tabcolsep}{3pt}
    \renewcommand{\arraystretch}{0.9}
    \resizebox{\linewidth}{!}{
    \begin{tabular}{l@{}l@{}ccc@{}ccc@{}}
      \toprule
      & & \multicolumn{3}{c@{}}{CIFAR-10} & \multicolumn{3}{c@{}}{ImageNet-256} \\
      \cmidrule(lr){3-5} \cmidrule(lr){6-8}
      Metric & & $N{=}10$ & $N{=}50$ & $N{=}150$ & $N{=}10$ & $N{=}50$ & $N{=}150$ \\
      \midrule
      $\rho~\downarrow$
      & \multirow{2}{*}{\rule{0pt}{1.1em}\textnormal{k-NN}}
      & $-0.54$ & $-0.61$ & $-0.65$ & $-0.38$ & $-0.42$ & $-0.38$ \\
      $\delta~\downarrow$
      &
      & $-0.83$ & $-0.89$ & $-0.93$ & $-0.55$ & $-0.58$ & $-0.55$ \\
      \midrule
      $\rho~\downarrow$
      & \multirow{2}{*}{\rule{0pt}{1.1em}\textnormal{KDE}}
      & $-0.54$ & $-0.61$ & $-0.64$ & $-0.31$ & $-0.33$ & $-0.31$ \\
      $\delta~\downarrow$
      &
      & $-0.82$ & $-0.88$ & $-0.92$ & $-0.43$ & $-0.47$ & $-0.43$ \\
      \bottomrule
    \end{tabular}}
    \captionof{table}{Correlation metrics of $k$-NN and KDE methods on CIFAR-10 and ImageNet-256 for different $N$.}
    \label{tab:knn_kde_correlation}
  \end{minipage}
  \hfill
  \begin{minipage}[c]{0.3\linewidth}
    \centering
    \includegraphics[width=\linewidth,clip,trim=0cm 0.2cm 0cm 1.2cm]{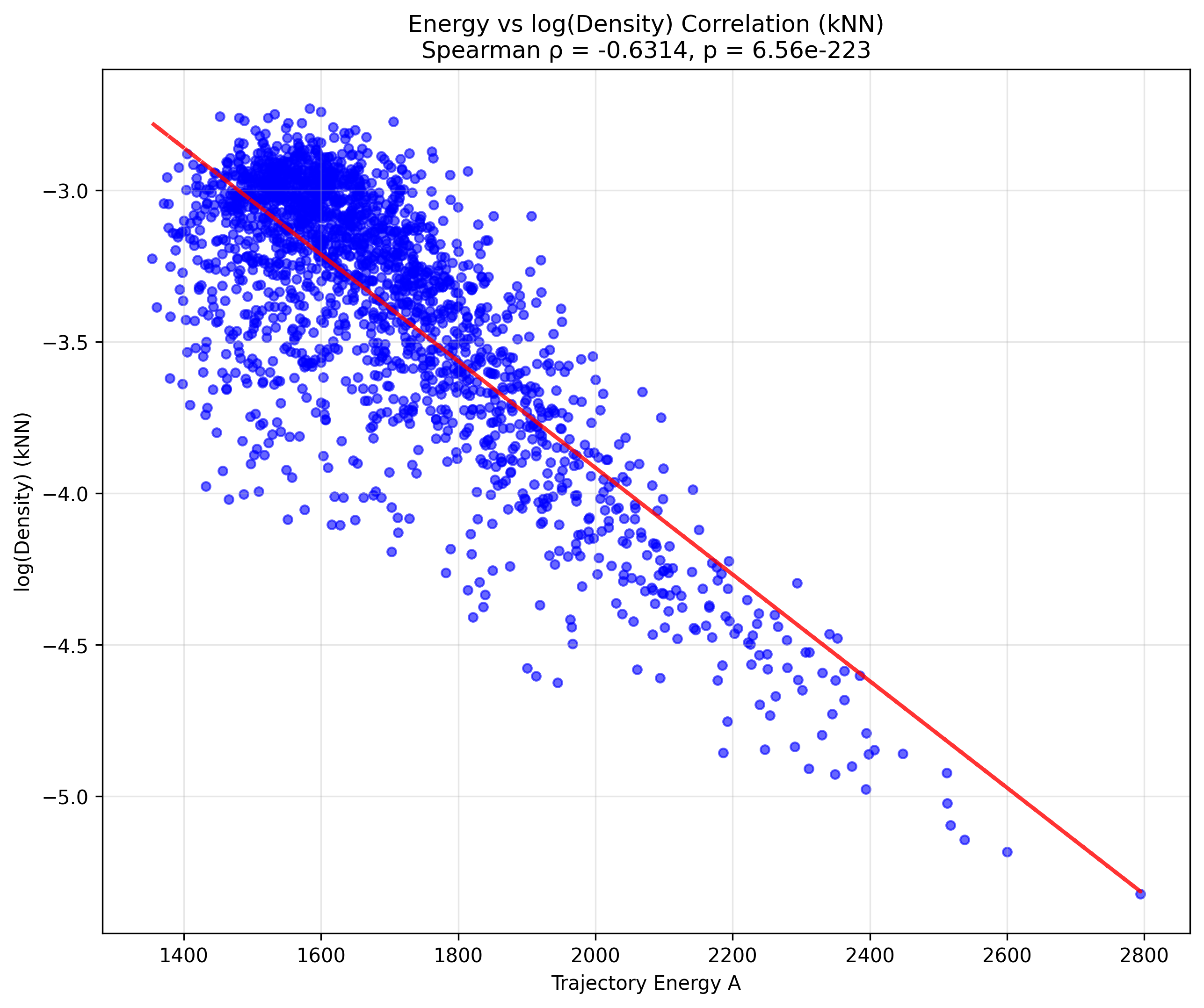}
    \captionsetup{width=1\linewidth}
    \captionof{figure}{\small $E$ vs.\ k-NN log-density ($\rho=-0.6314$, $p=6.56\times10^{-223}$).}
    \label{fig:energy_density_scatter_knn}
  \end{minipage}
  \vspace{-3.3mm}
\end{figure}

\textbf{Quantitative Results.} 
Table \ref{tab:knn_kde_correlation} shows  \SL{consistent} negative correlations between trajectory energy and data density. For CIFAR-10, Spearman’s $\rho$ drops from $-0.54$ to $-0.65$ (k-NN) and $-0.54$ to $-0.64$ (KDE) as $N$ increases, with similar trends for Cliff’s $\delta$ and ImageNet-256. 
\autoref{fig:energy_density_scatter_knn} shows a \SL{scatter plot of $E$ vs. k-NN's log-density}, with blue dots as samples and a red trend line. 
More visualizations and detailed statistics are provided in the Appendix.

\textbf{Conclusion.}
\emph{\SL{ ${E}$ inversely correlates with the estimated data densities, suggesting} that high-energy trajectories \SL{tend to concentrate in sparse regions of the data manifold.}}

\section{Conclusion}
\label{sec:conclusion}
\vspace{-3mm}
\SL{We introduced Kinetic Path Energy (KPE), a physics-inspired diagnostic measuring the kinetic cost of flow-based generation. Our experiments reveal that higher KPE correlates with stronger semantic quality and lower data density, showing that informative samples naturally lie in sparse regions. This trajectory-level view offers interpretable insights beyond traditional endpoint-based evaluation, and motivates future theoretical work on the connection between KPE and generative dynamics, as well as extensions  to stochastic samplers through the lens of stochastic thermodynamics \cite{seifert2012stochastic, ikeda2025speed}.}

\begin{ack}
This work was supported by the Wallenberg AI, Autonomous Systems and Software Program (WASP) funded by the Knut and Alice Wallenberg Foundation. We acknowledge the computational resources provided by the Alvis cluster.
SHL would like to acknowledge support from the Wallenberg Initiative on Networks and Quantum Information (WINQ) and the Swedish Research Council (VR/2021-03648)
\end{ack}
\bibliographystyle{unsrtnat}
\bibliography{enfopath}

\newpage
\section{Appendix}
\label{sec:appendix}

\subsection{Additional Experiments on Kinetic Path Energy (KPE) vs. Semantic Strength}

\paragraph{Qualitative Examples}
We provide qualitative visual comparisons across diverse ImageNet-256 classes to demonstrate the consistent semantic quality differences between high-energy and low-energy trajectories.

\begin{figure}[htbp]
    \vspace{-3mm}
    \centering
    \includegraphics[width=0.8\textwidth]{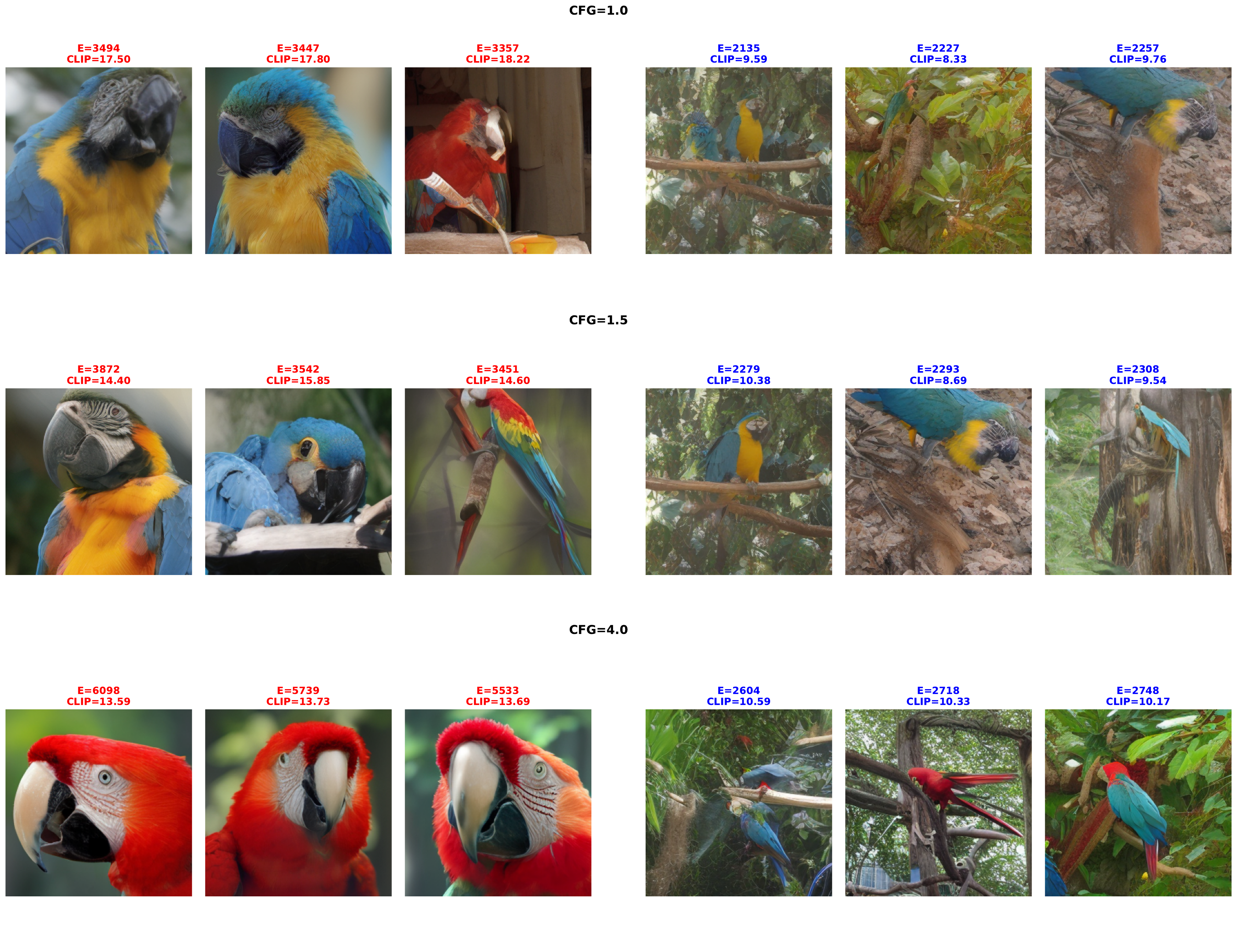}
    \caption{\textbf{Macaw} (ImageNet-256): High-KPE (left) vs. low-KPE (right) across CFG scales 1.0, 1.5, 4.0. Higher KPE yields richer semantic details, vibrant colors, and sharper textures.}
    \label{fig:semantic_vis_macaw}
\end{figure}

\begin{figure}[htbp]
    \vspace{-3mm}
    \centering
    \includegraphics[width=0.8\textwidth]{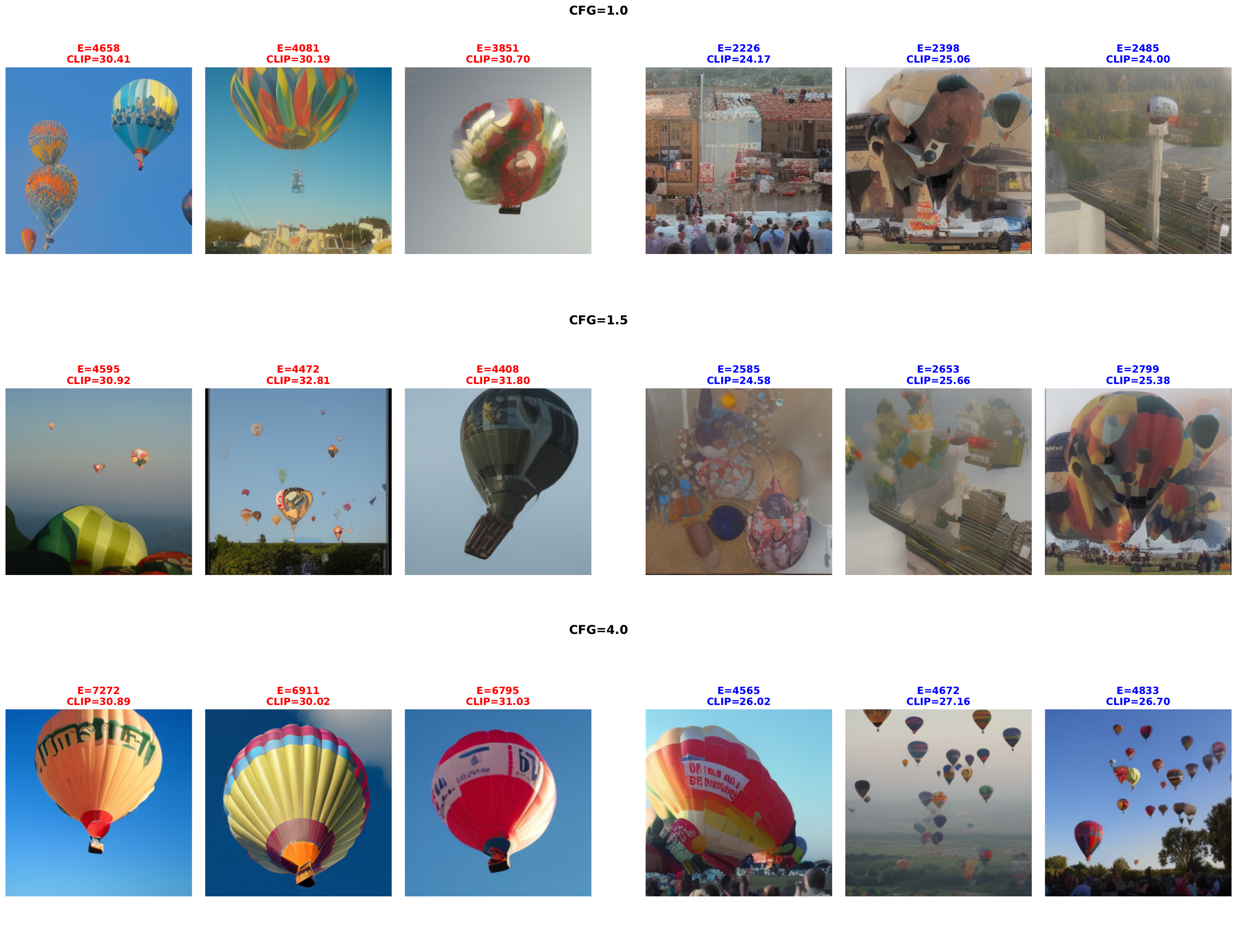}
    \caption{\textbf{Hot Air Balloon} (ImageNet-256): High-KPE (left) vs. low-KPE (right) across CFG scales 1.0, 1.5, 4.0. Higher KPE shows clearer structures and better color saturation.}
    \label{fig:semantic_vis_balloon}
\end{figure}

\begin{figure}[htbp]
    \centering
    \includegraphics[width=0.8\textwidth]{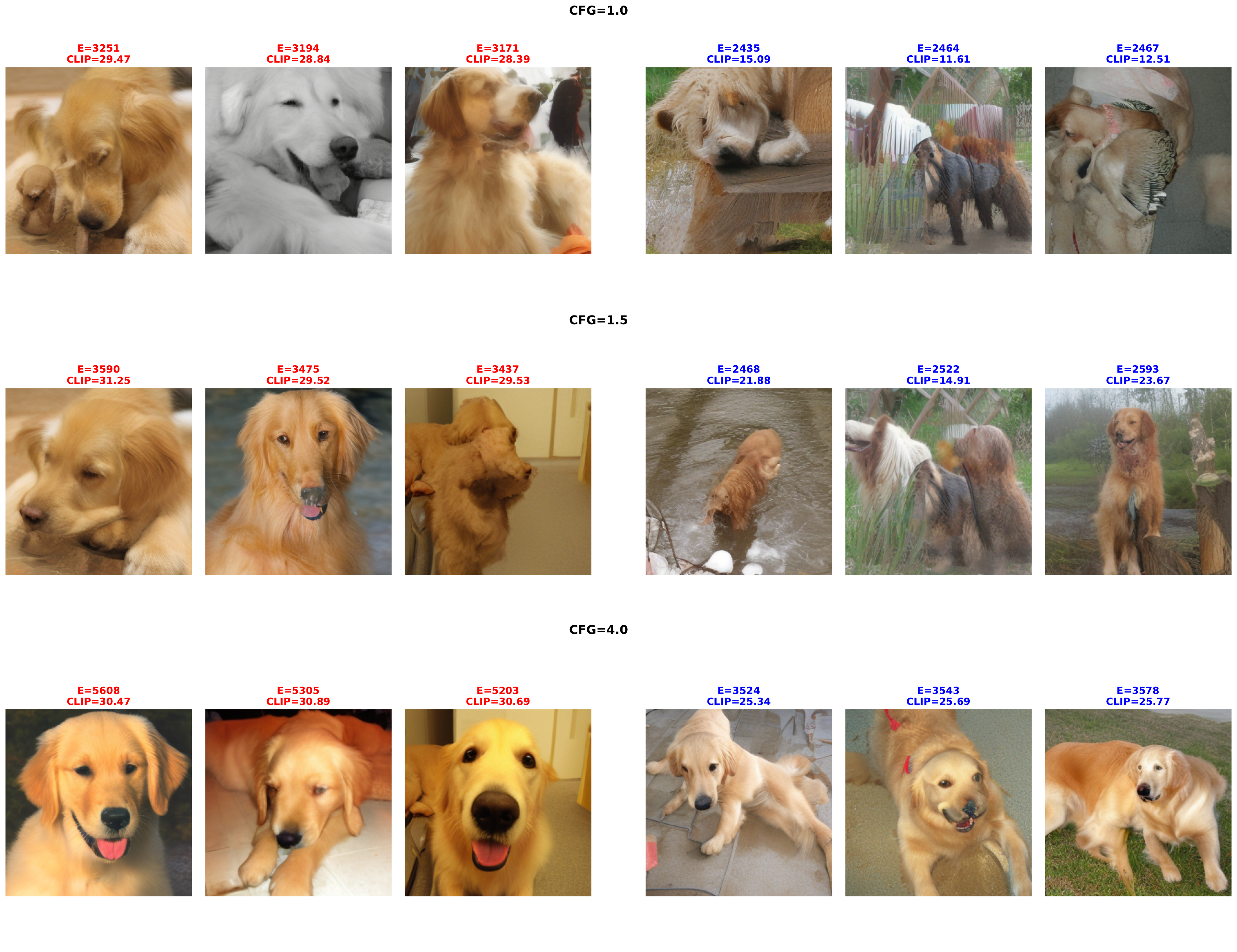}
    \caption{\textbf{Golden Retriever} (ImageNet-256): High-KPE (left) vs. low-KPE (right) across CFG scales 1.0, 1.5, 4.0. Higher KPE produces finer textures and clearer facial features.}
    \label{fig:semantic_vis_golden_retriever}
\end{figure}

\begin{figure}[htbp]
    \centering
    \includegraphics[width=0.8\textwidth]{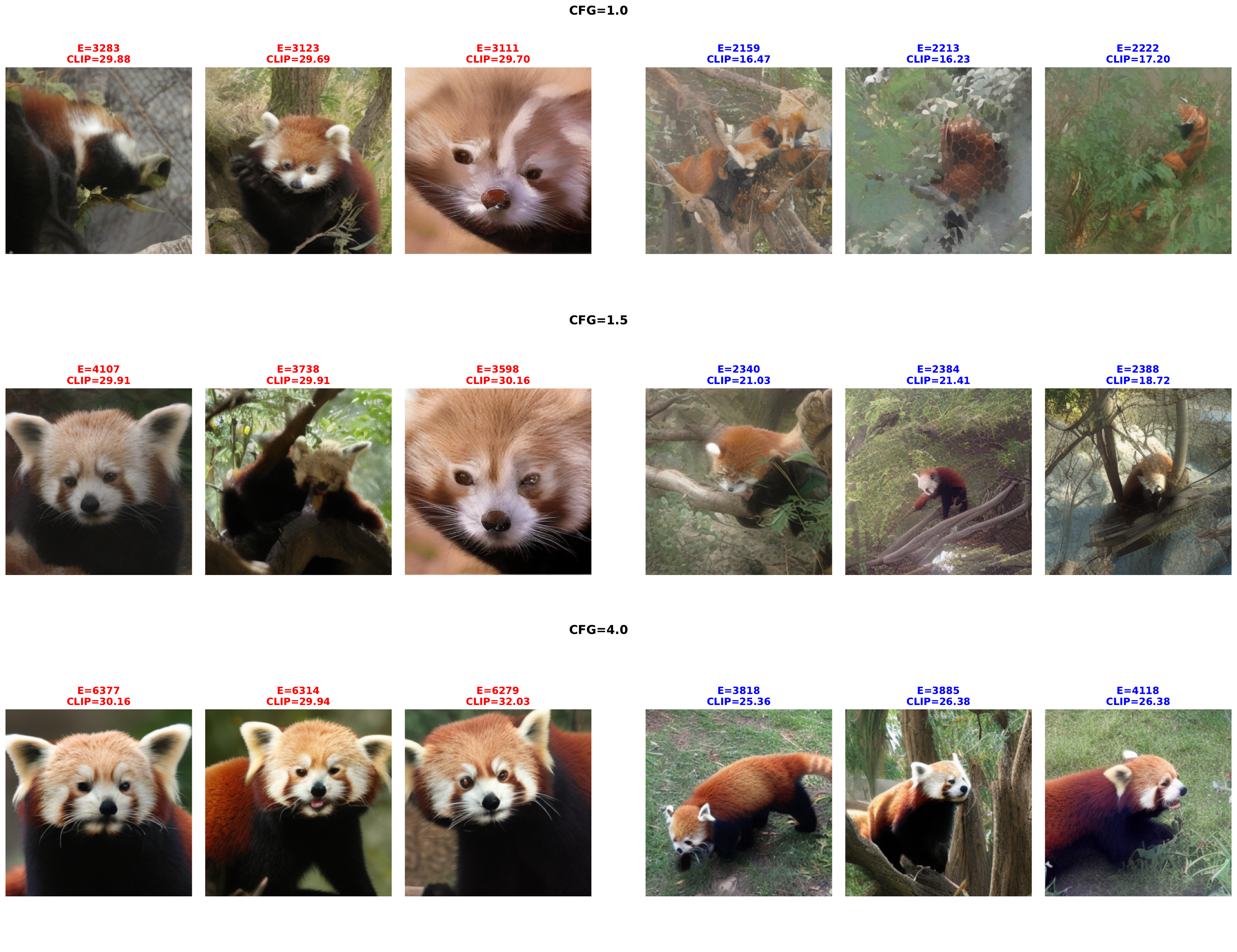}
    \caption{\textbf{African Elephant} (ImageNet-256): High-KPE (left) vs. low-KPE (right) across CFG scales 1.0, 1.5, 4.0. Higher KPE shows more defined features and better skin texture.}
    \label{fig:semantic_vis_african_elephant}
\end{figure}

\begin{figure}[htbp]
    \centering
    \includegraphics[width=0.8\textwidth]{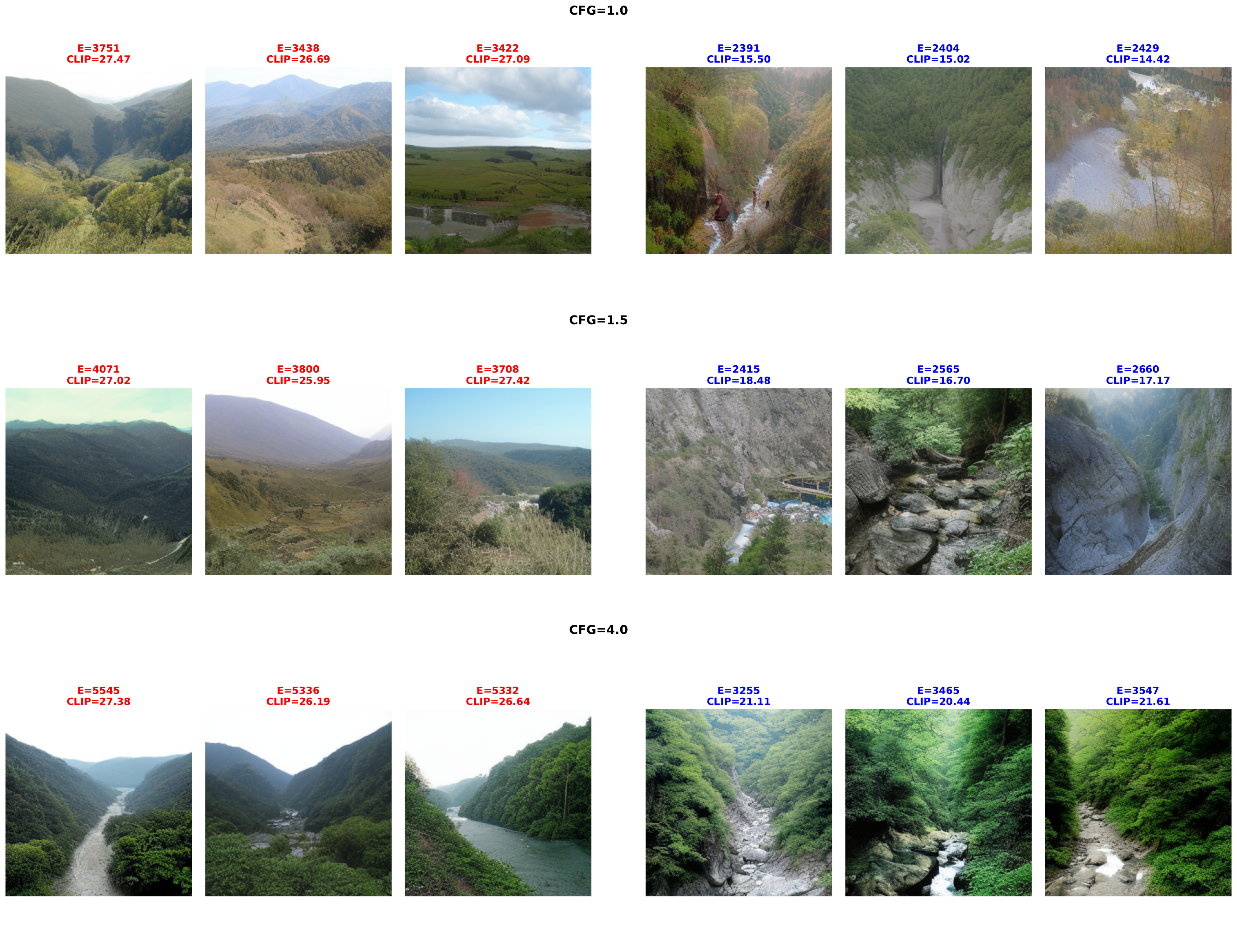}
    \caption{\textbf{Valley} (ImageNet-256): High-KPE (left) vs. low-KPE (right) across CFG scales 1.0, 1.5, 4.0. Higher KPE generates more detailed terrain and better depth perception.}
    \label{fig:semantic_vis_valley}
\end{figure}

\begin{figure}[htbp]
    \centering
    \includegraphics[width=0.8\textwidth]{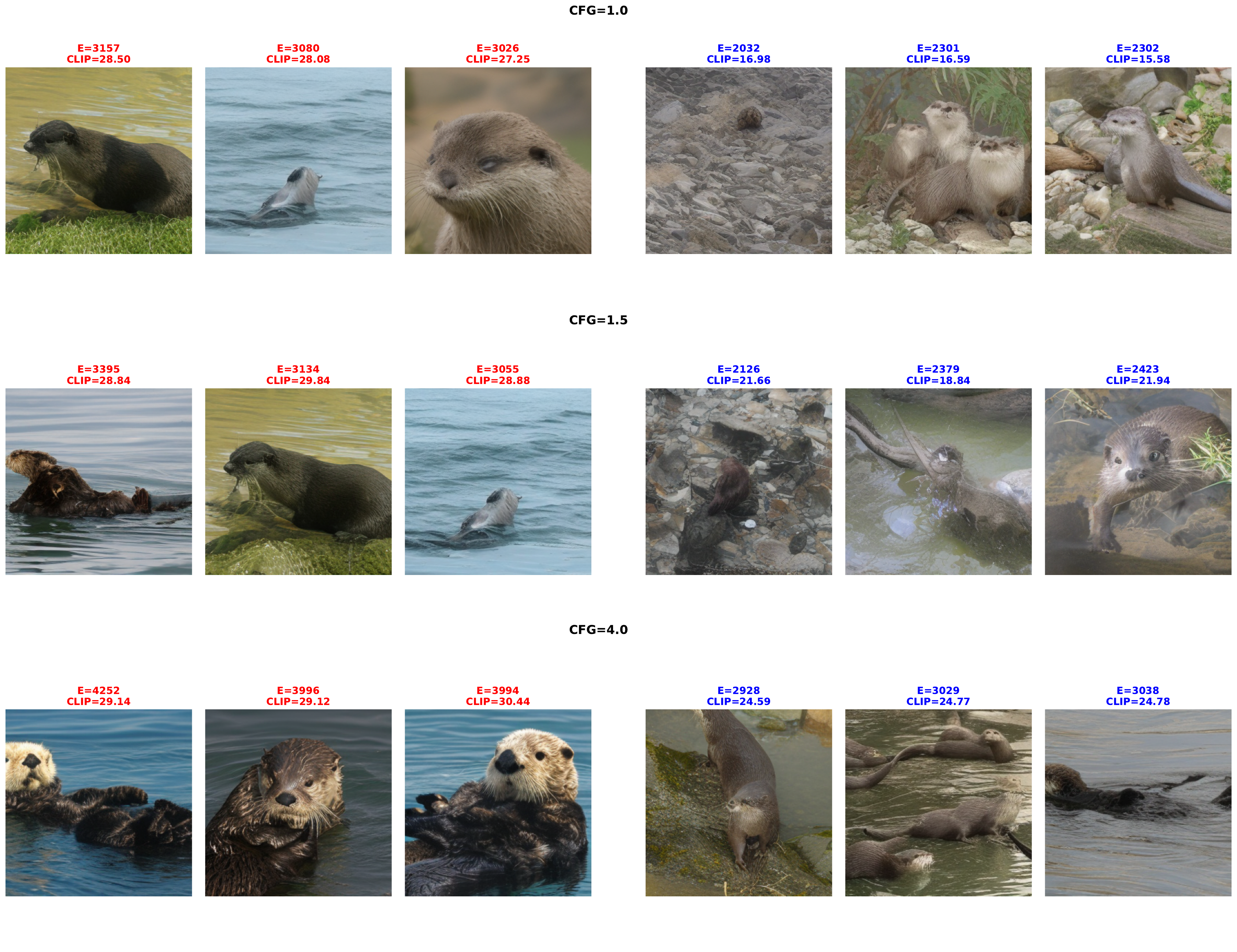}
    \caption{\textbf{Otter} (ImageNet-256): High-KPE (left) vs. low-KPE (right) across CFG scales 1.0, 1.5, 4.0. Higher KPE shows sharper outlines and more realistic details.}
    \label{fig:semantic_vis_otter}
\end{figure}

\newpage
\subsection{Additional Experiments on Kinetic Path Energy (KPE) vs. Data Density}
This section provides comprehensive evidence supporting our second key finding: higher KPE strongly correlates with lower data density. We present correlation analysis, effect size quantification, spatial distribution patterns, and robustness validation across different sampling configurations.

\begin{figure}[htbp]
    \centering
    \begin{subfigure}{0.48\textwidth}
        \centering
        \includegraphics[width=\textwidth]{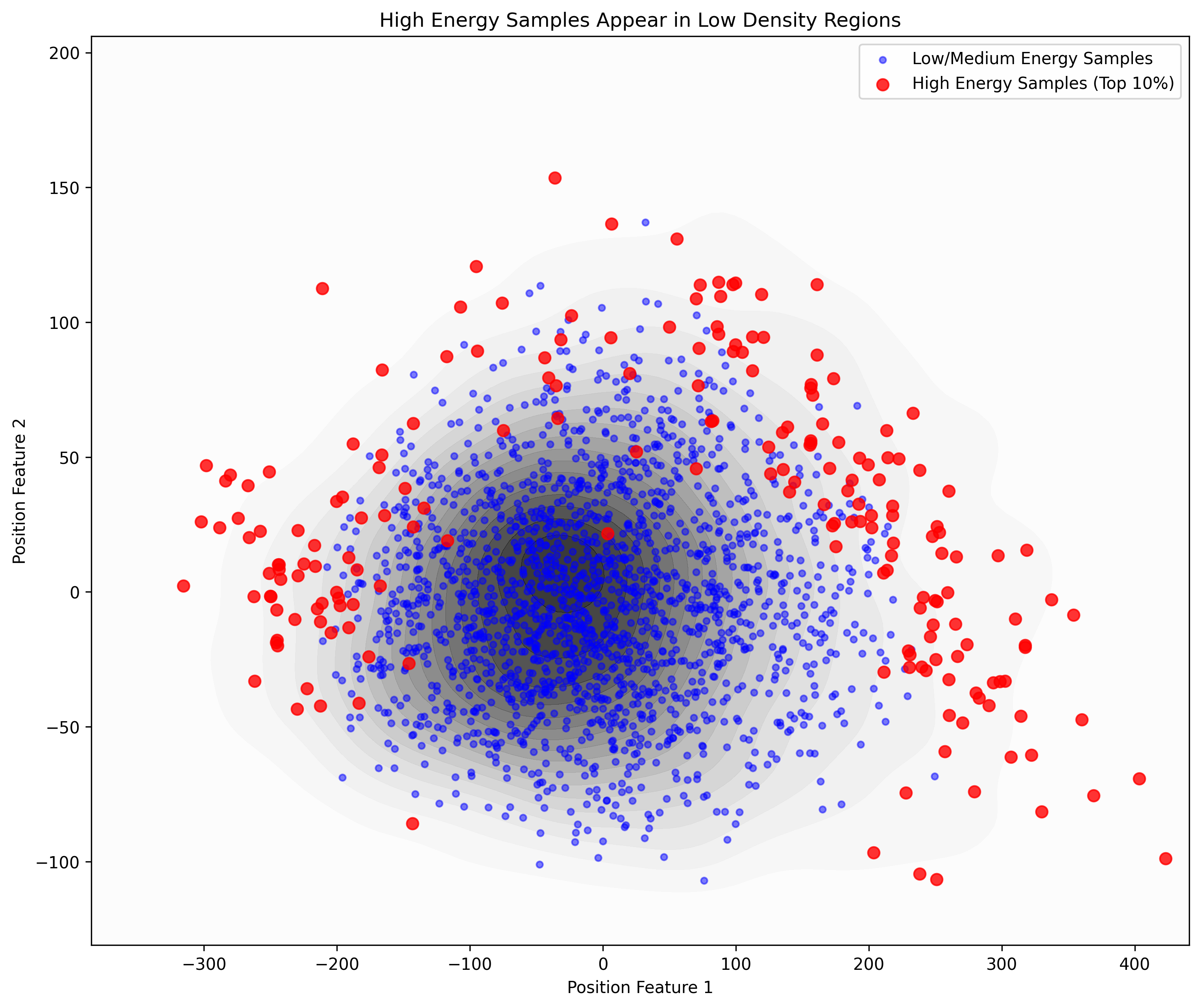}
        \caption{10 sampling steps}
        \label{fig:high_energy_10steps}
    \end{subfigure}
    \hfill
    \begin{subfigure}{0.48\textwidth}
        \centering
        \includegraphics[width=\textwidth]{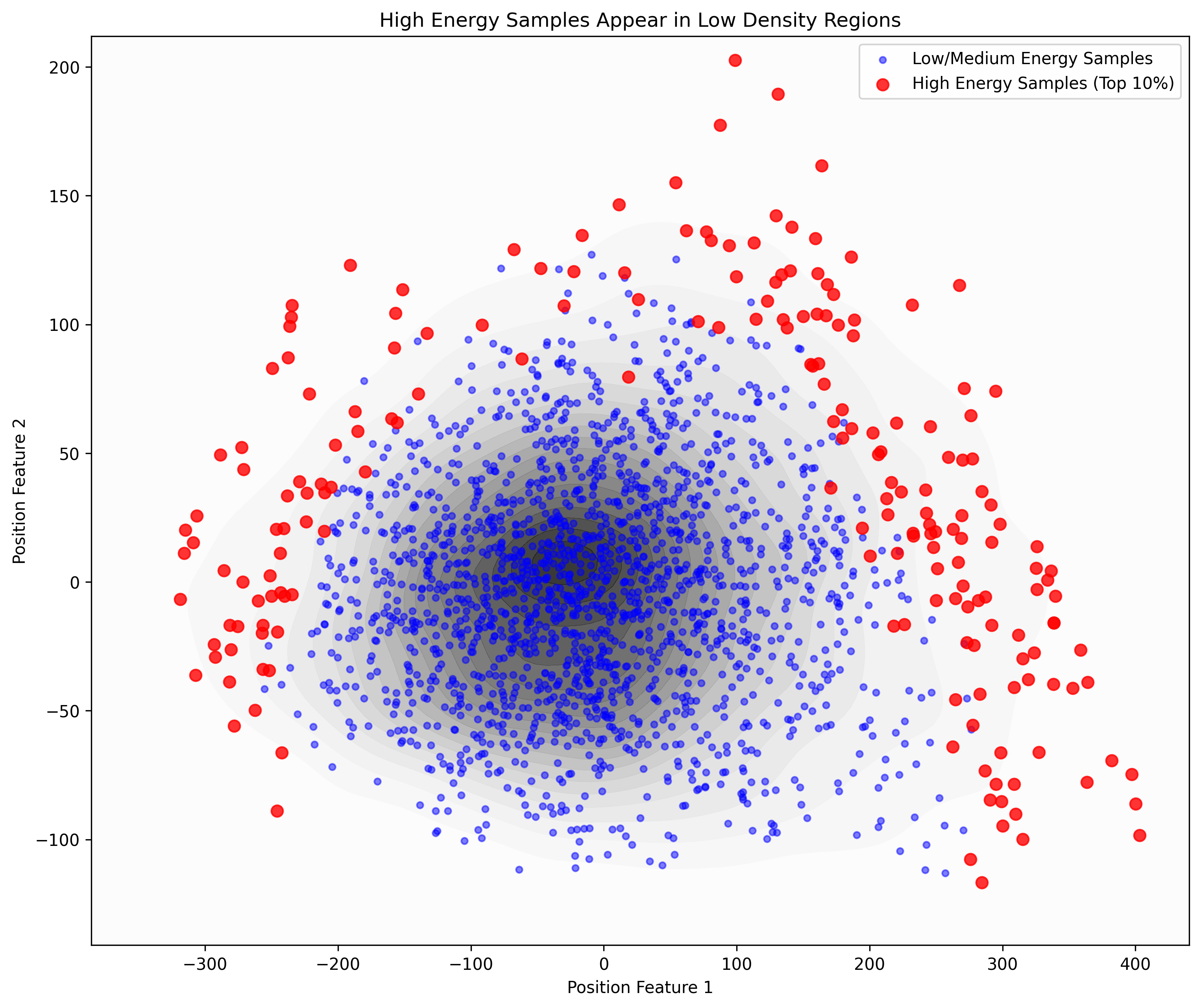}
        \caption{50 sampling steps}
        \label{fig:high_energy_50steps}
    \end{subfigure}

    \vspace{0.5cm}

    \begin{subfigure}{0.48\textwidth}
        \centering
        \includegraphics[width=\textwidth]{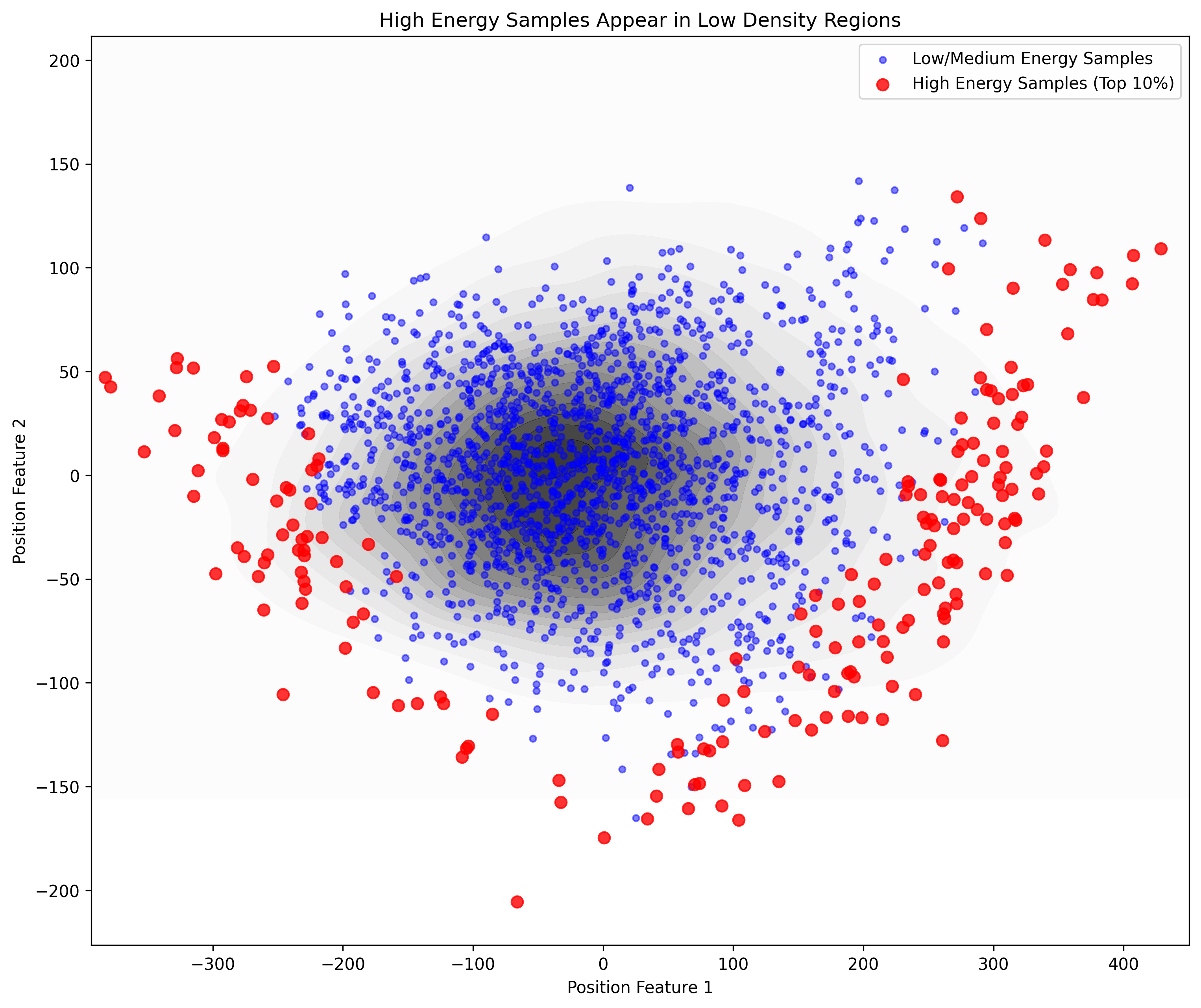}
        \caption{100 sampling steps}
        \label{fig:high_energy_100steps}
    \end{subfigure}
    \hfill
    \begin{subfigure}{0.48\textwidth}
        \centering
        \includegraphics[width=\textwidth]{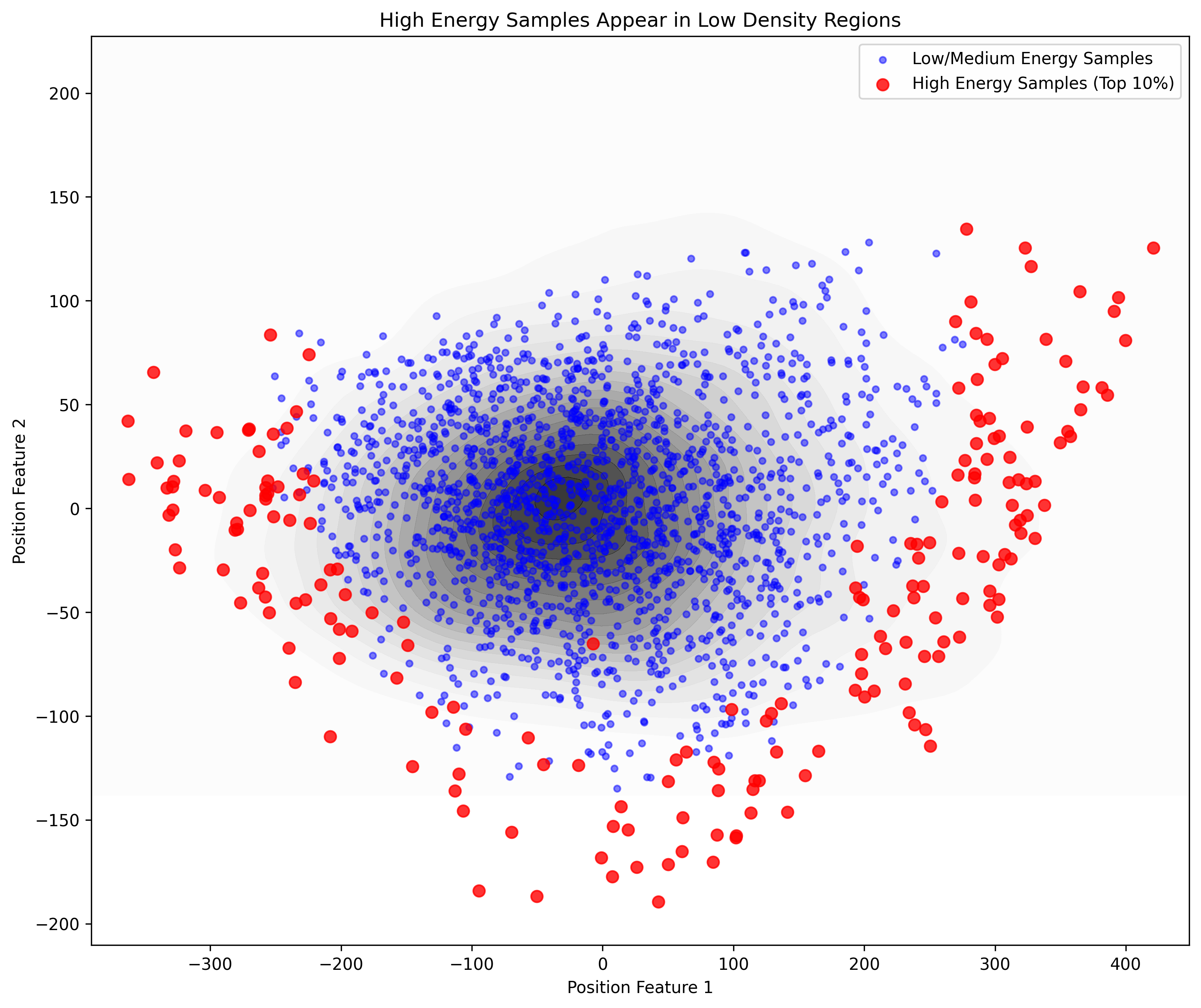}
        \caption{150 sampling steps}
        \label{fig:high_energy_150steps}
    \end{subfigure}

    \caption{2D visualization of kinetic path energy (KPE) and data density. Red points represent the top 10\% highest-energy samples, which consistently fall in low-density regions. Blue points represent low/medium-energy samples. Contours indicate density levels, highlighting the inverse correlation between KPE and density.}
    \label{fig:high_energy_spatial}
\end{figure}

\begin{figure}[htbp]
    \centering
    \begin{subfigure}{0.4\textwidth}
        \centering
        \includegraphics[width=\textwidth]{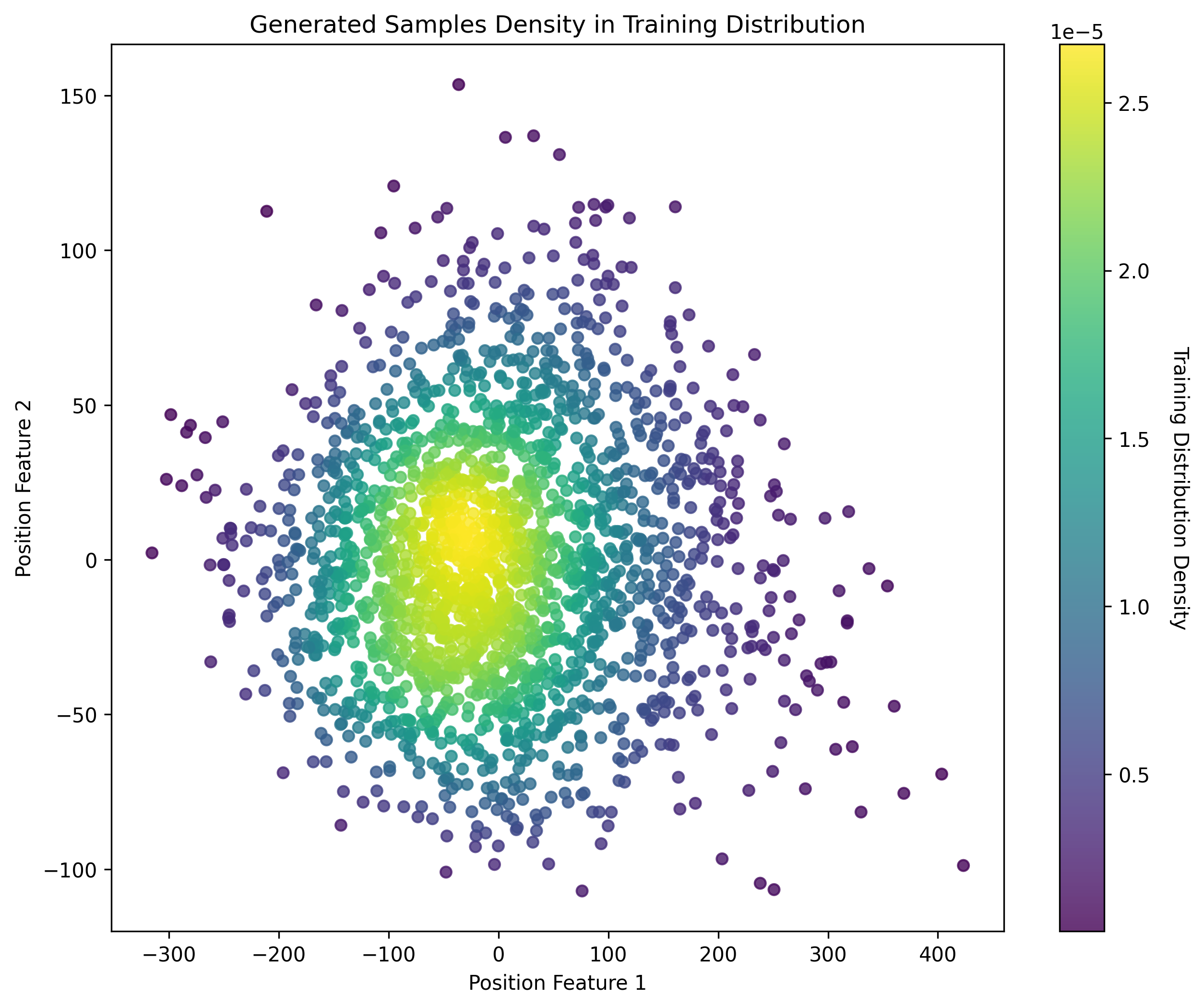}
        \caption{10 steps - Density distribution in 2D}
        \label{fig:gen_samples_density_10steps}
    \end{subfigure}
    \hfill
    \begin{subfigure}{0.4\textwidth}
        \centering
        \includegraphics[width=\textwidth]{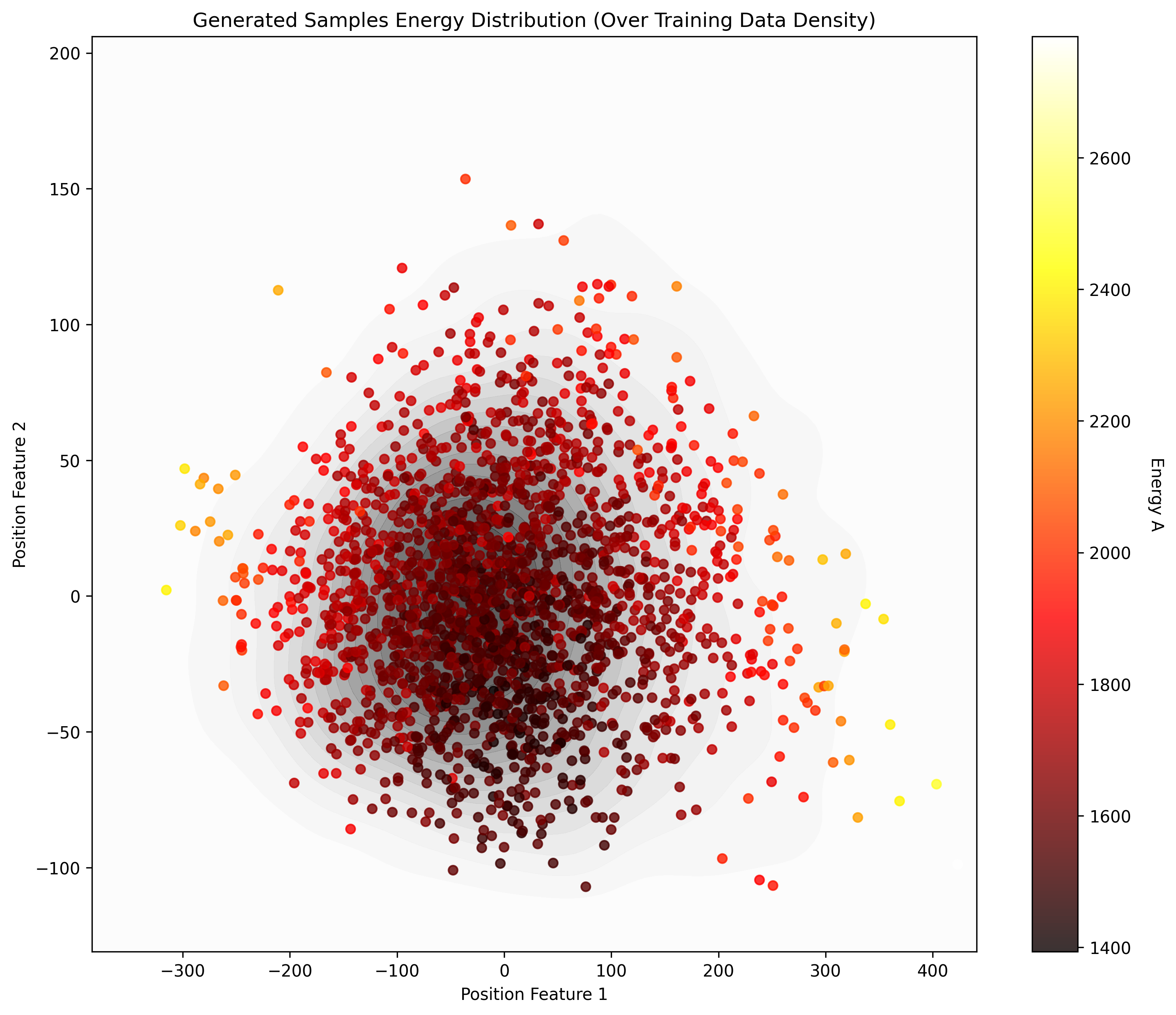}
        \caption{10 steps - KPE distribution in 2D}
        \label{fig:energy_contour_10steps}
    \end{subfigure}

    \begin{subfigure}{0.4\textwidth}
        \centering
        \includegraphics[width=\textwidth]{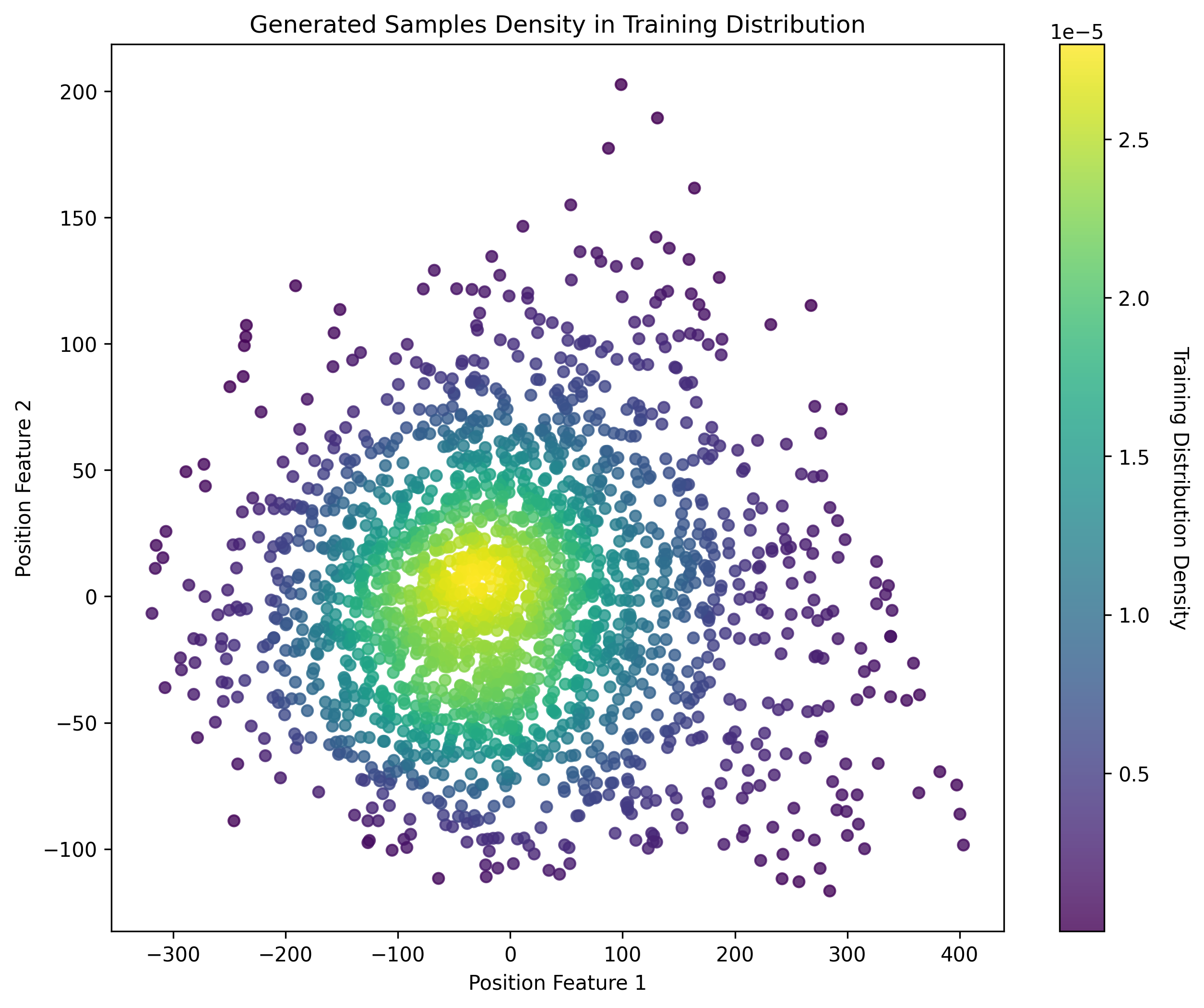}
        \caption{50 steps - Density distribution in 2D}
        \label{fig:gen_samples_density_50steps}
    \end{subfigure}
    \hfill
    \begin{subfigure}{0.4\textwidth}
        \centering
        \includegraphics[width=\textwidth]{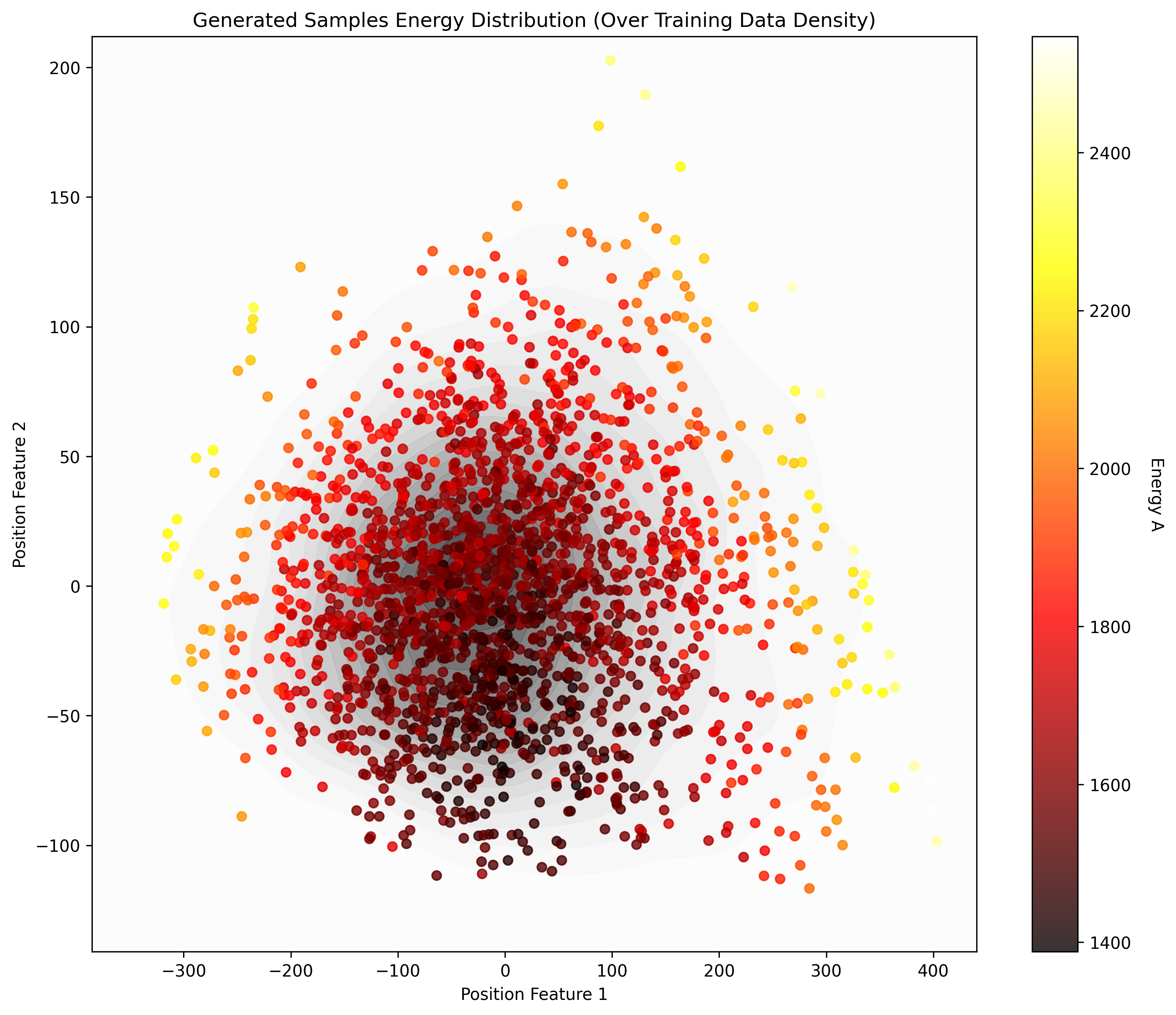}
        \caption{50 steps - KPE distribution in 2D}
        \label{fig:energy_contour_50steps}
    \end{subfigure}

    \begin{subfigure}{0.4\textwidth}
        \centering
        \includegraphics[width=\textwidth]{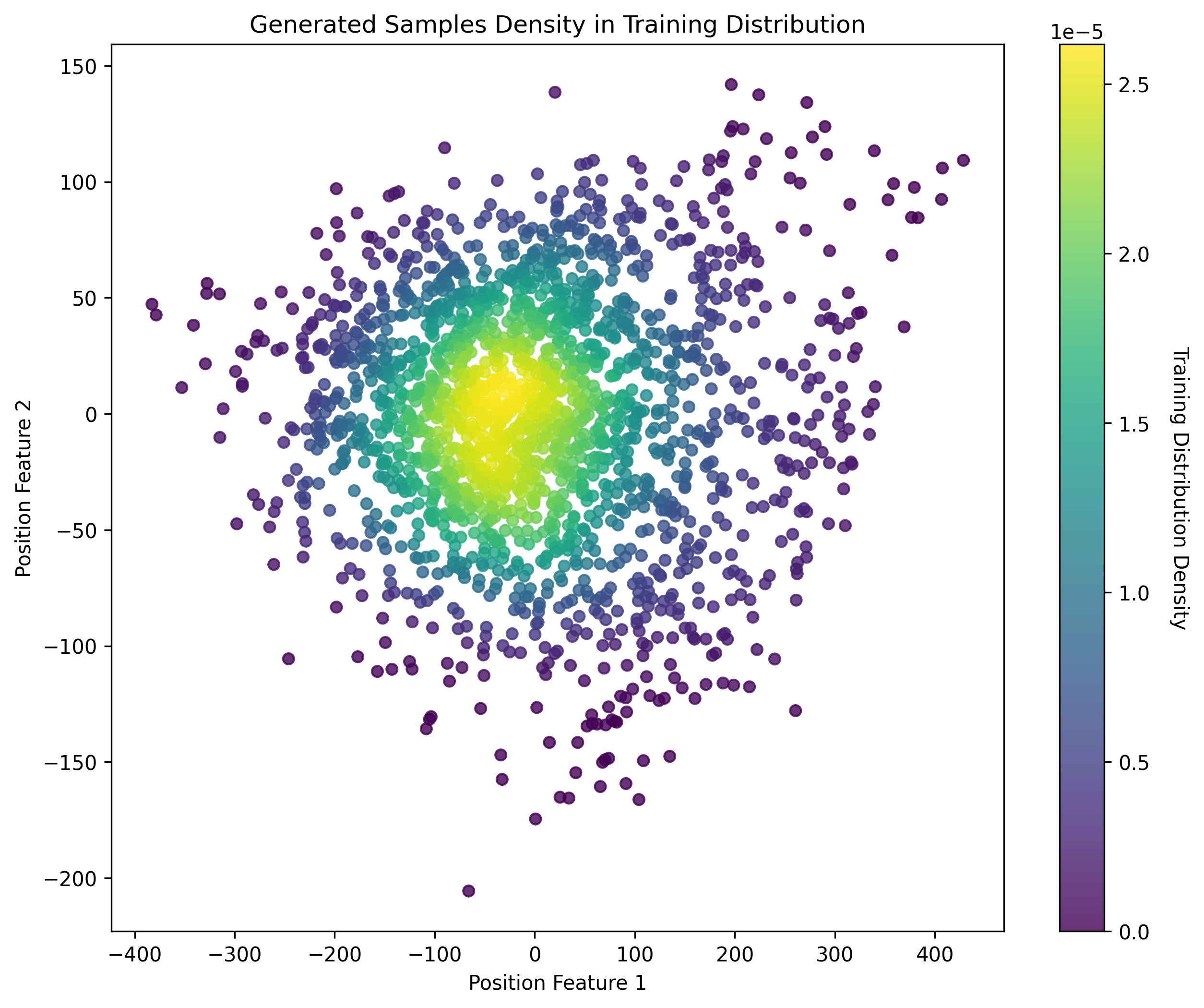}
        \caption{100 steps - Density distribution in 2D}
        \label{fig:gen_samples_density_100steps}
    \end{subfigure}
    \hfill
    \begin{subfigure}{0.4\textwidth}
        \centering
        \includegraphics[width=\textwidth]{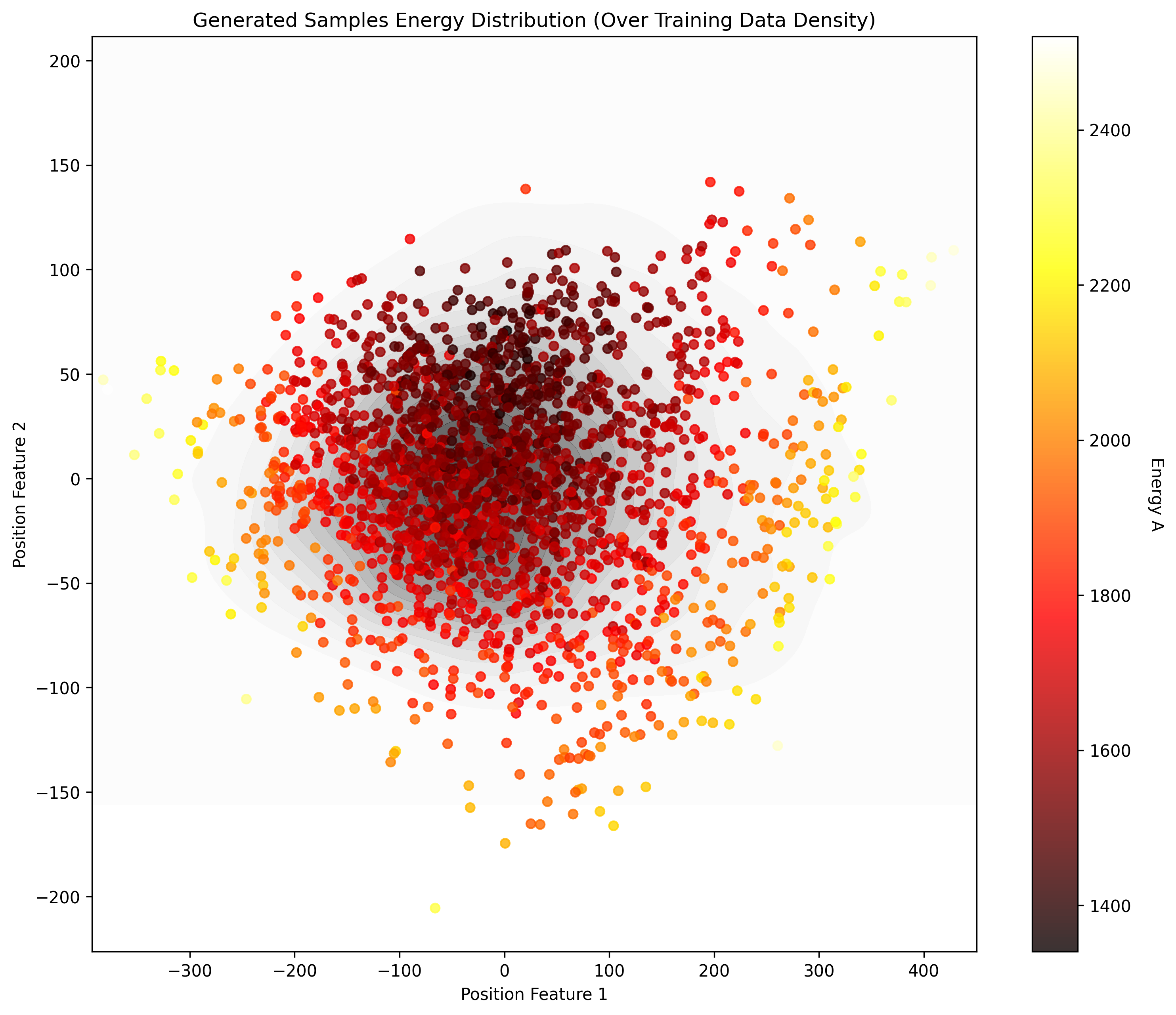}
        \caption{100 steps - KPE distribution in 2D}
        \label{fig:energy_contour_100steps}
    \end{subfigure}

    \begin{subfigure}{0.4\textwidth}
        \centering
        \includegraphics[width=\textwidth]{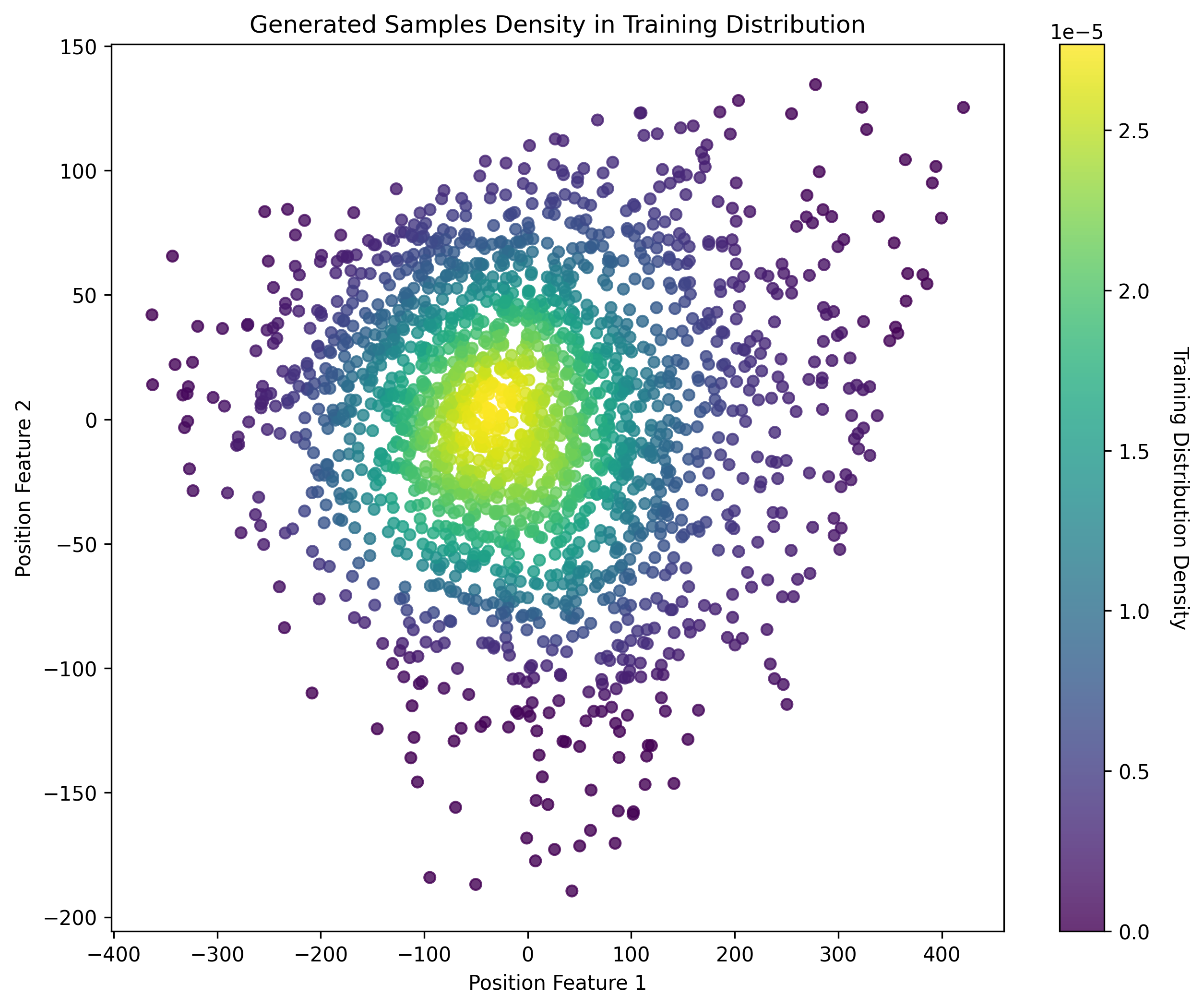}
        \caption{150 steps - Density distribution in 2D}
        \label{fig:gen_samples_density_150steps}
    \end{subfigure}
    \hfill
    \begin{subfigure}{0.4\textwidth}
        \centering
        \includegraphics[width=\textwidth]{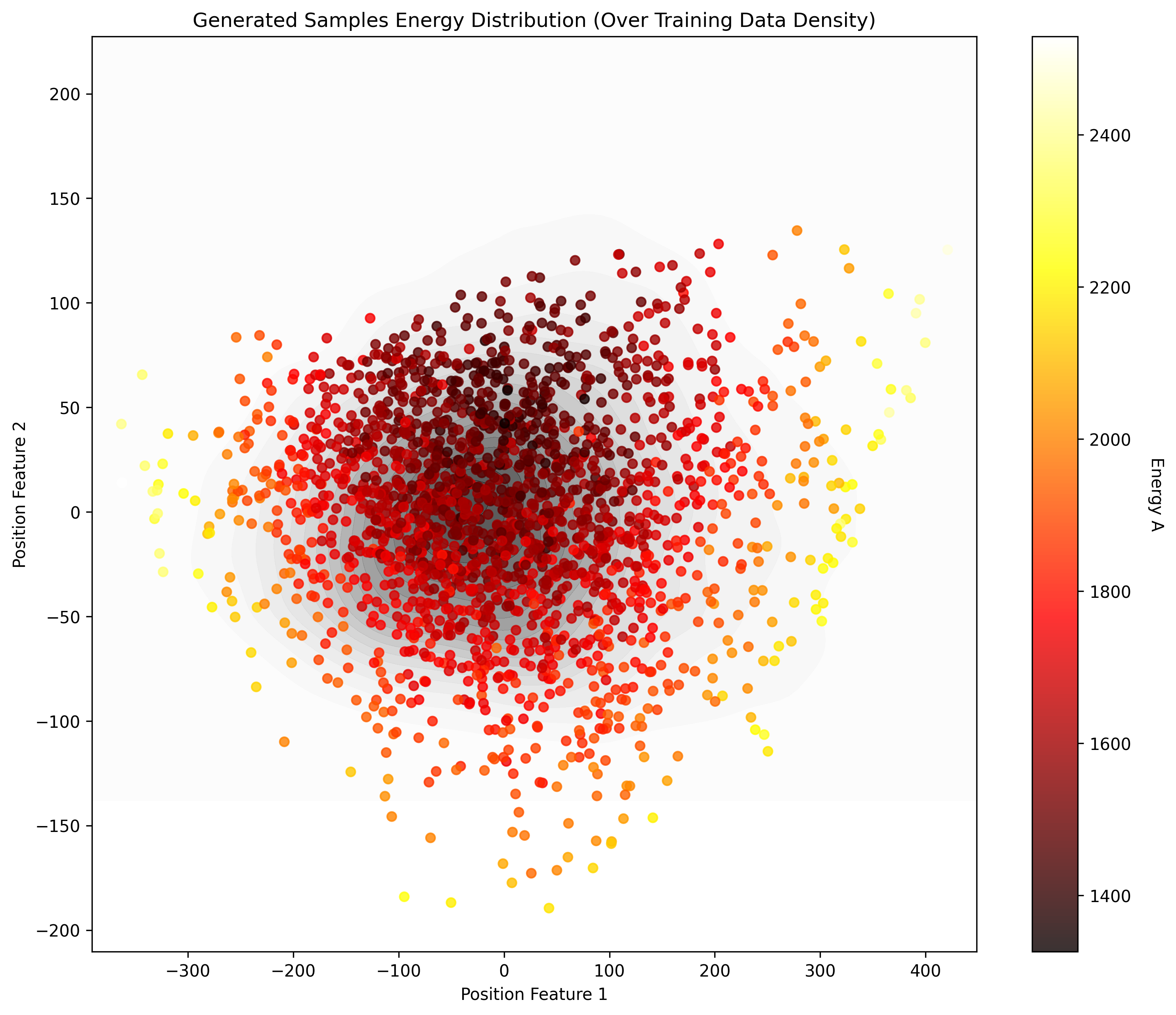}
        \caption{150 steps - KPE distribution in 2D}
        \label{fig:energy_contour_150steps}
    \end{subfigure}

    \caption{2D visualizations. Left column shows the distribution-density visualization of generated samples projected to 2D using PCA. Right column presents the corresponding kinetic path energy (KPE) in 2D across different sampling steps (10, 50, 100, 150 rows). }
    \label{fig:energy_density_additional}
\end{figure}

\begin{figure}[htbp]
    \centering
    \begin{subfigure}{0.48\textwidth}
        \centering
        \includegraphics[width=\textwidth]{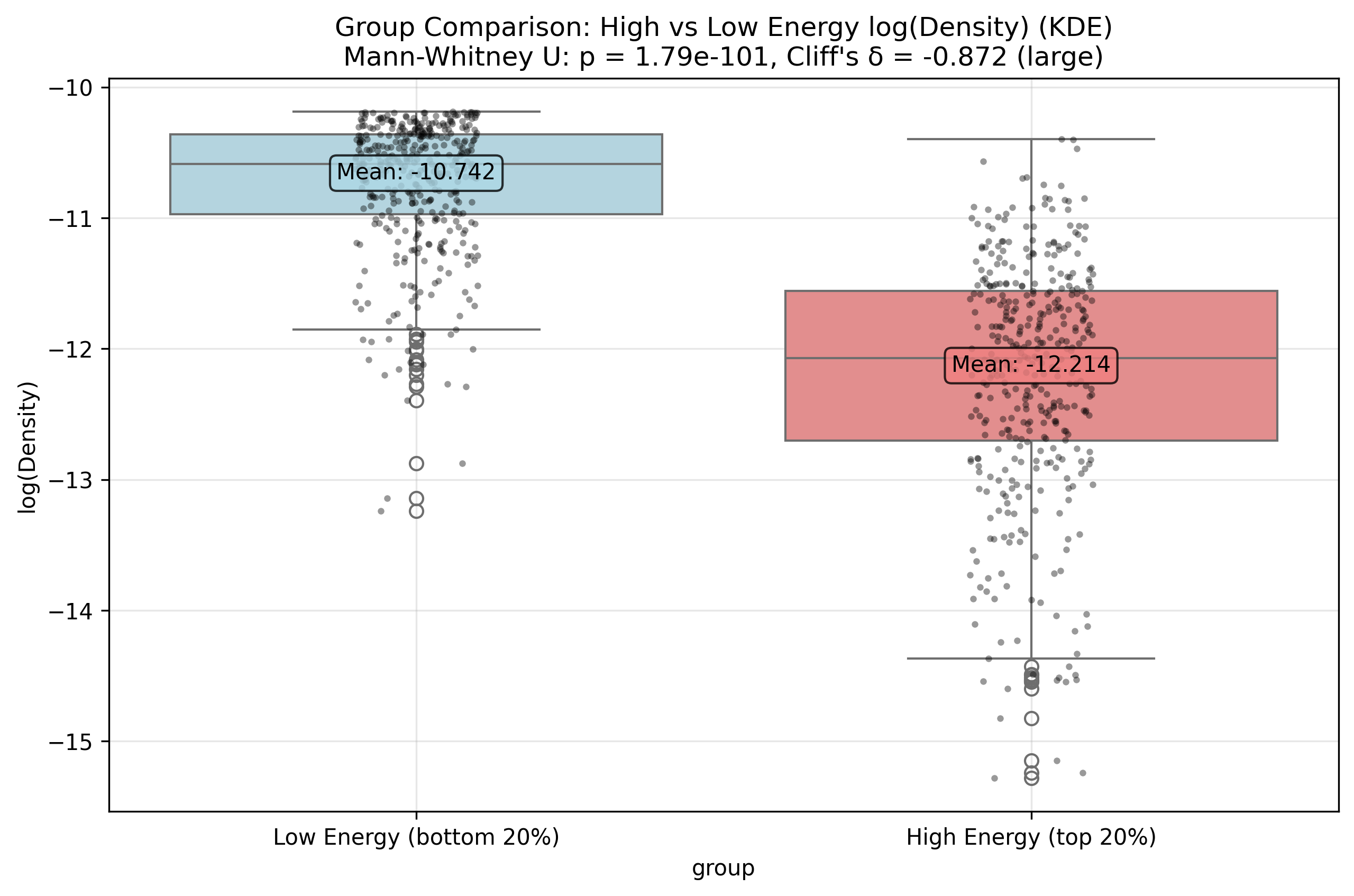}
        \caption{10 steps - KDE ($\delta = -0.872$)}
        \label{fig:group_comparison_10steps_kde}
    \end{subfigure}
    \hfill
    \begin{subfigure}{0.48\textwidth}
        \centering
        \includegraphics[width=\textwidth]{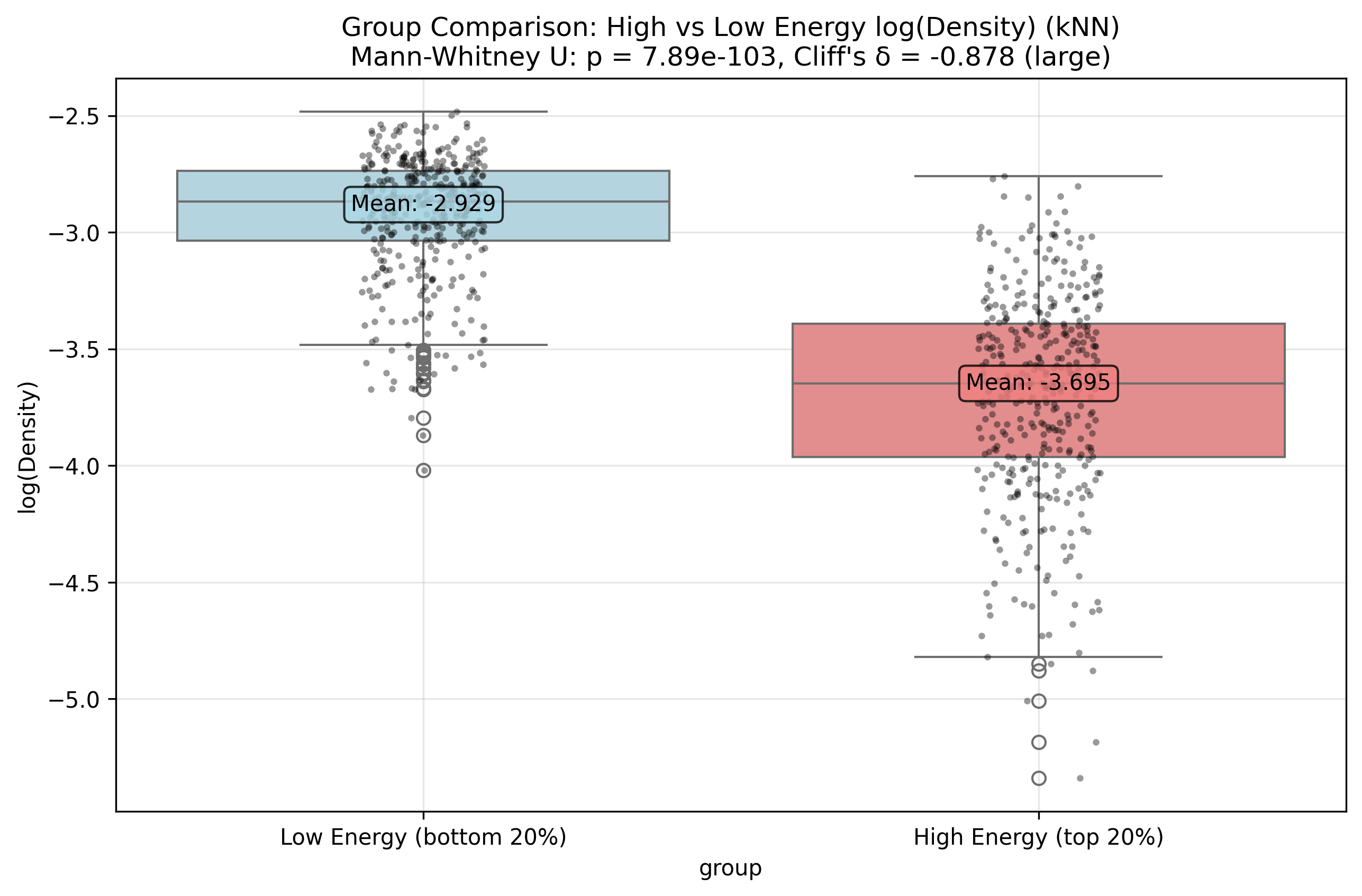}
        \caption{10 steps - k-NN ($\delta = -0.878$)}
        \label{fig:group_comparison_10steps_knn}
    \end{subfigure}

    \vspace{0.3cm}

    \begin{subfigure}{0.48\textwidth}
        \centering
        \includegraphics[width=\textwidth]{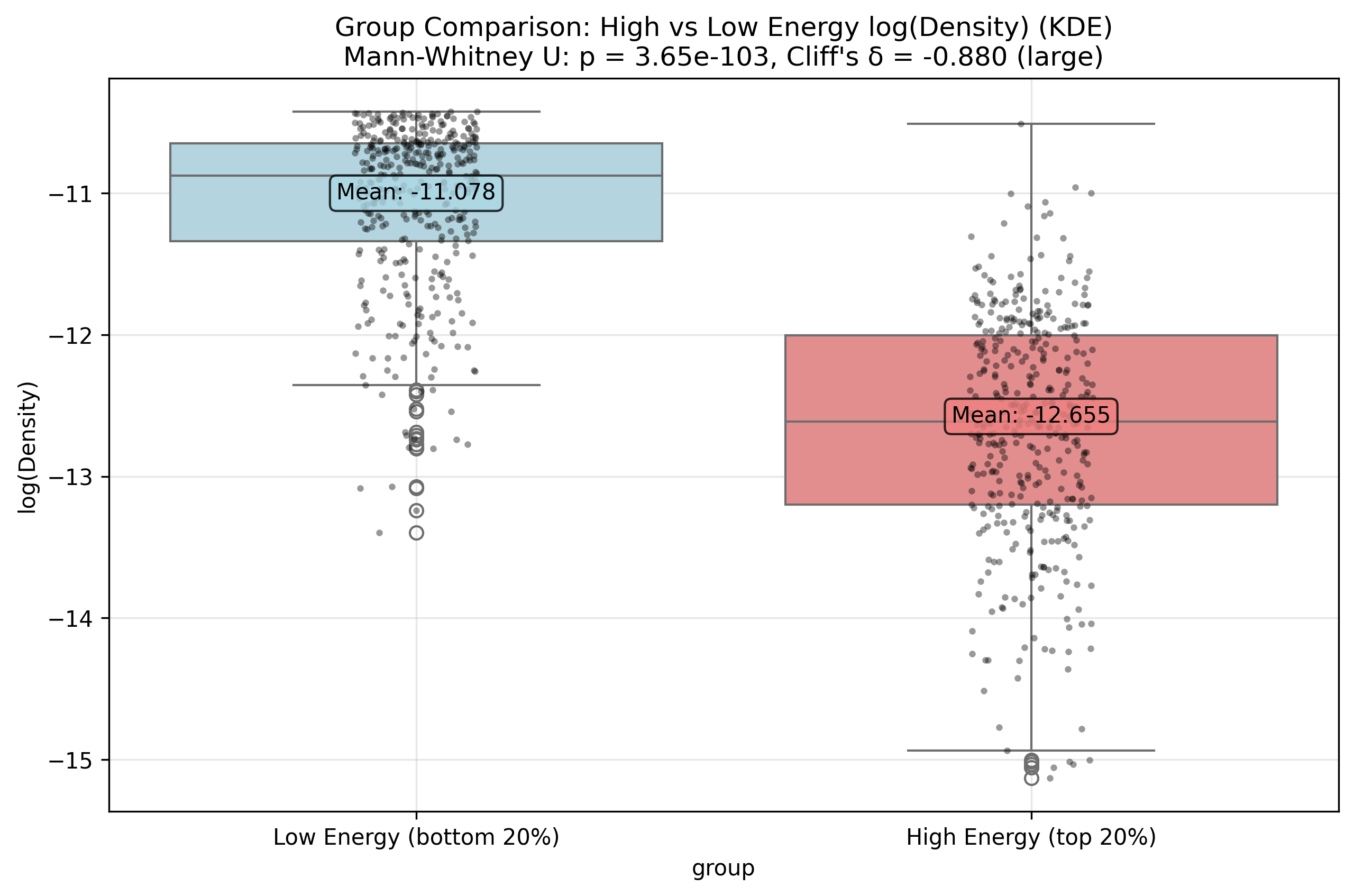}
        \caption{50 steps - KDE ($\delta = -0.880$)}
        \label{fig:group_comparison_50steps_kde}
    \end{subfigure}
    \hfill
    \begin{subfigure}{0.48\textwidth}
        \centering
        \includegraphics[width=\textwidth]{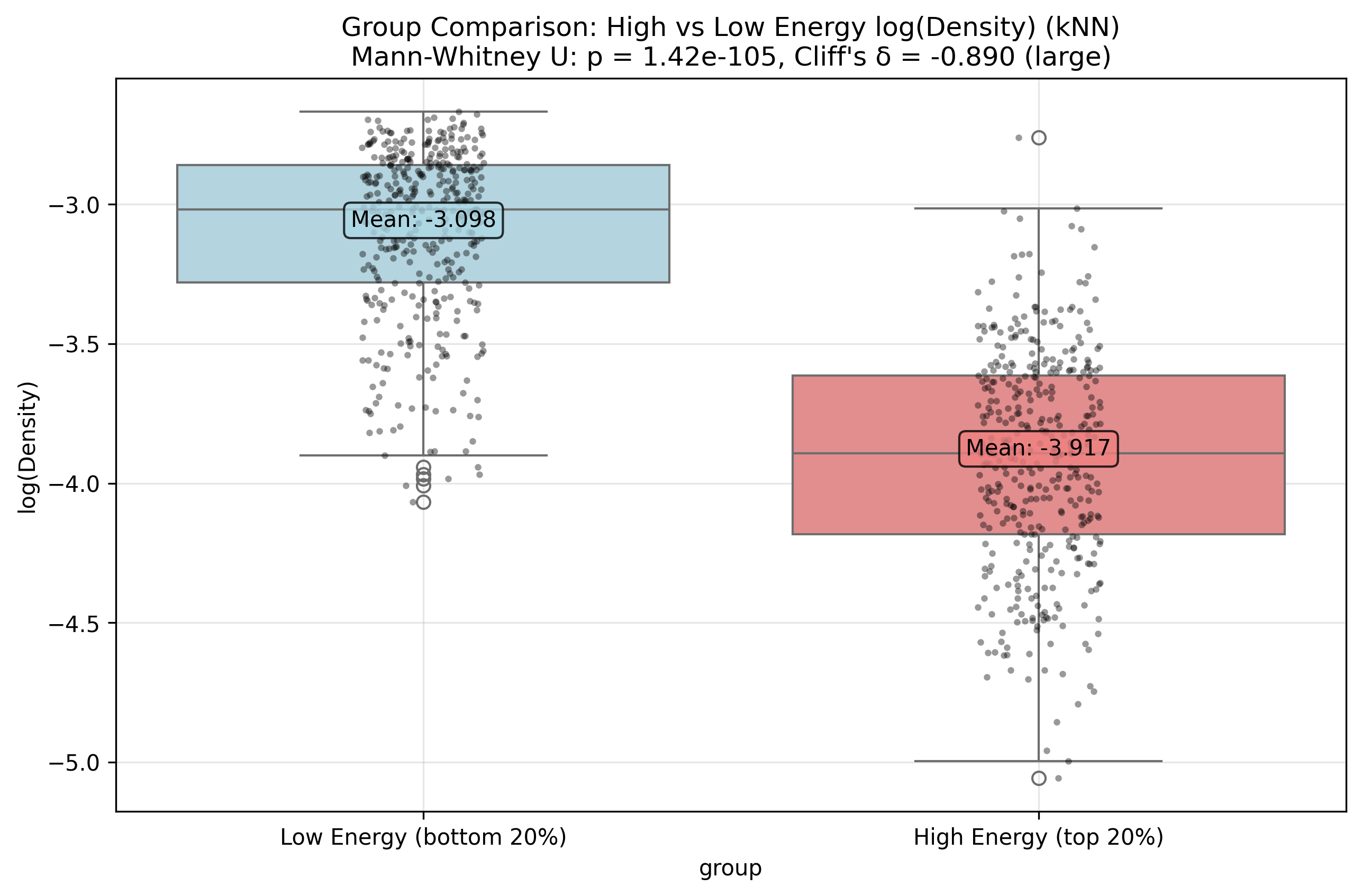}
        \caption{50 steps - k-NN ($\delta = -0.890$)}
        \label{fig:group_comparison_50steps_knn}
    \end{subfigure}

    \vspace{0.3cm}

    \begin{subfigure}{0.48\textwidth}
        \centering
        \includegraphics[width=\textwidth]{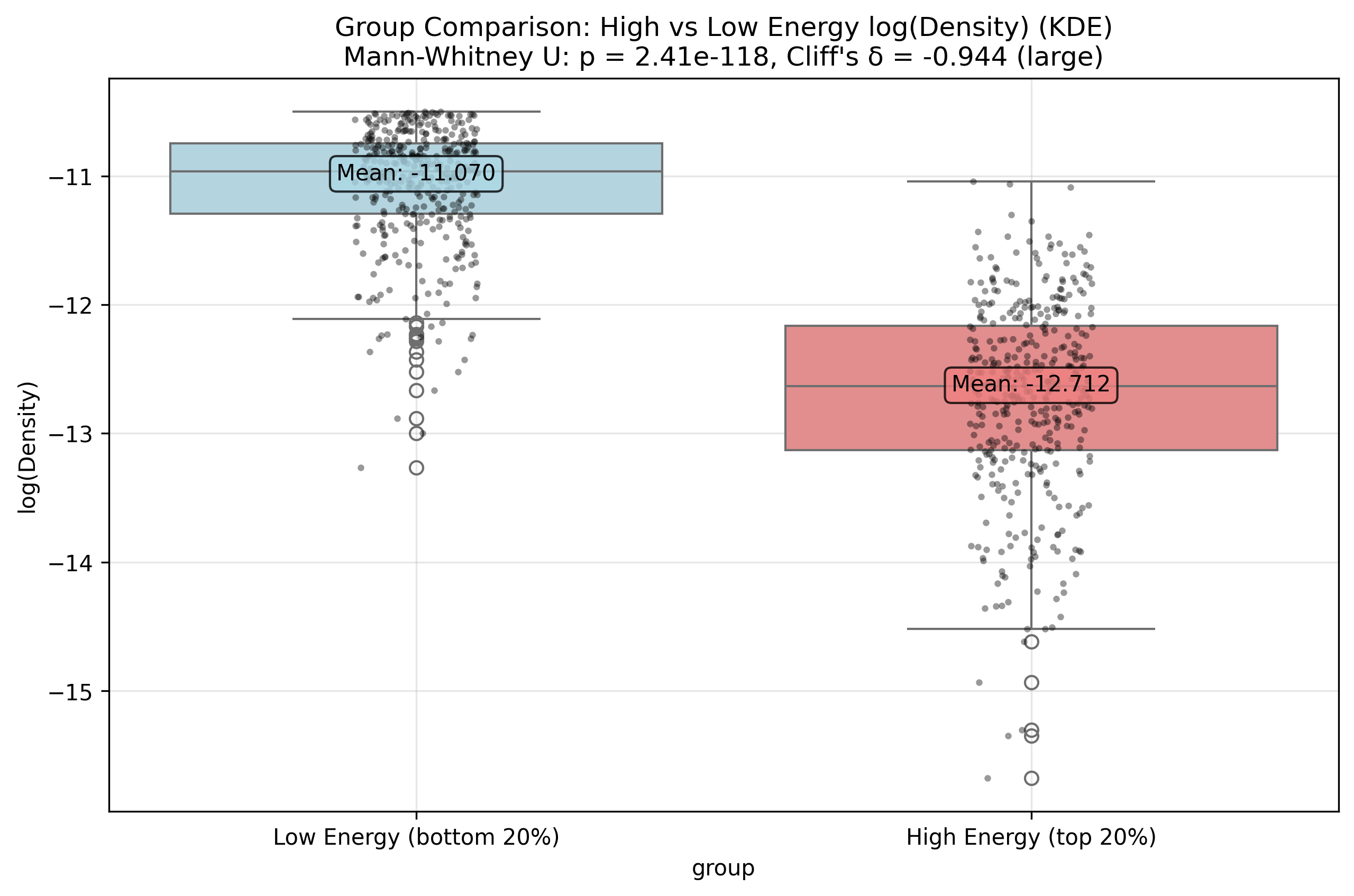}
        \caption{100 steps - KDE ($\delta = -0.944$)}
        \label{fig:group_comparison_100steps_kde}
    \end{subfigure}
    \hfill
    \begin{subfigure}{0.48\textwidth}
        \centering
        \includegraphics[width=\textwidth]{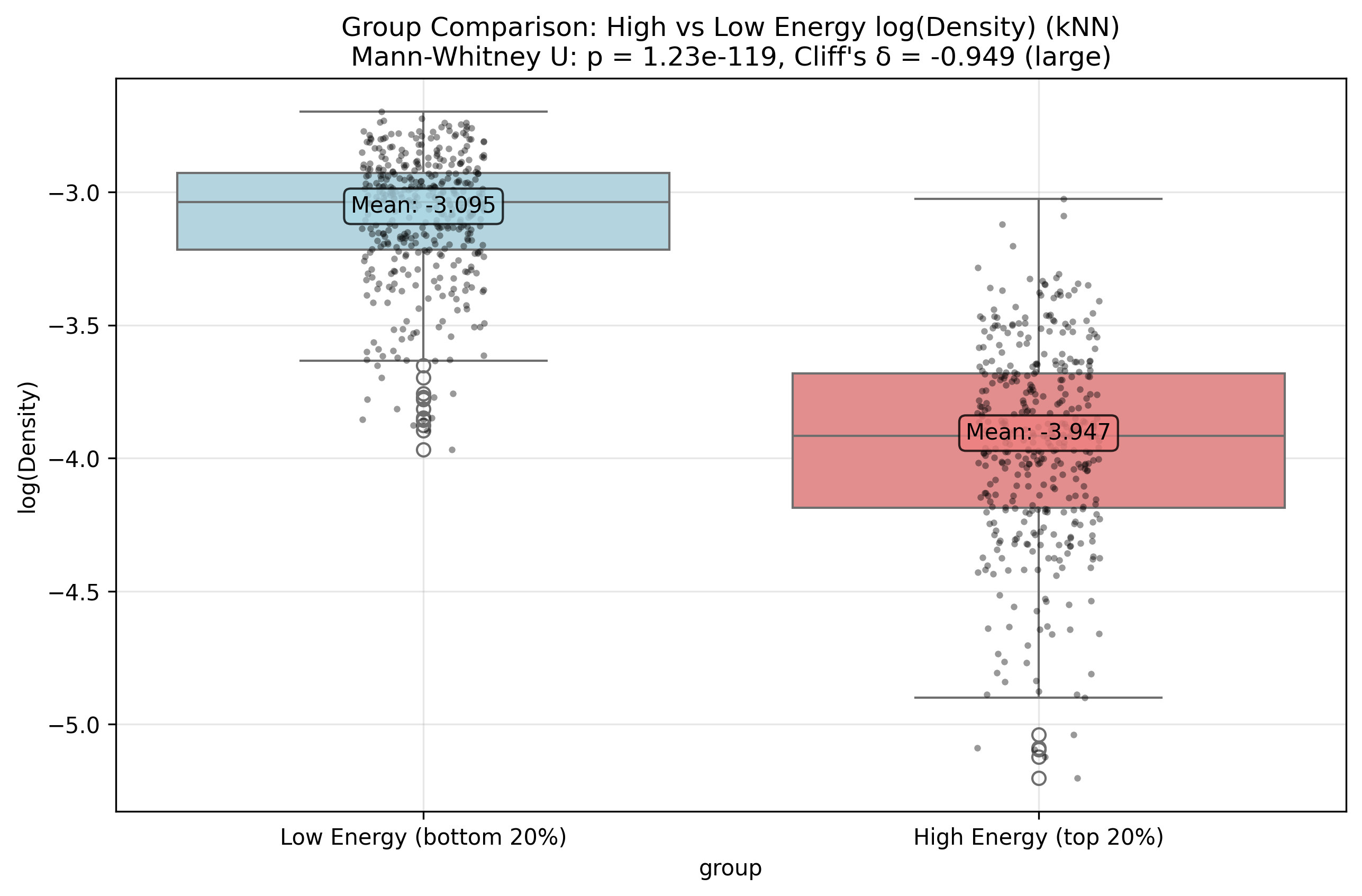}
        \caption{100 steps - k-NN ($\delta = -0.949$)}
        \label{fig:group_comparison_100steps_knn}
    \end{subfigure}

    \vspace{0.3cm}

    \begin{subfigure}{0.48\textwidth}
        \centering
        \includegraphics[width=\textwidth]{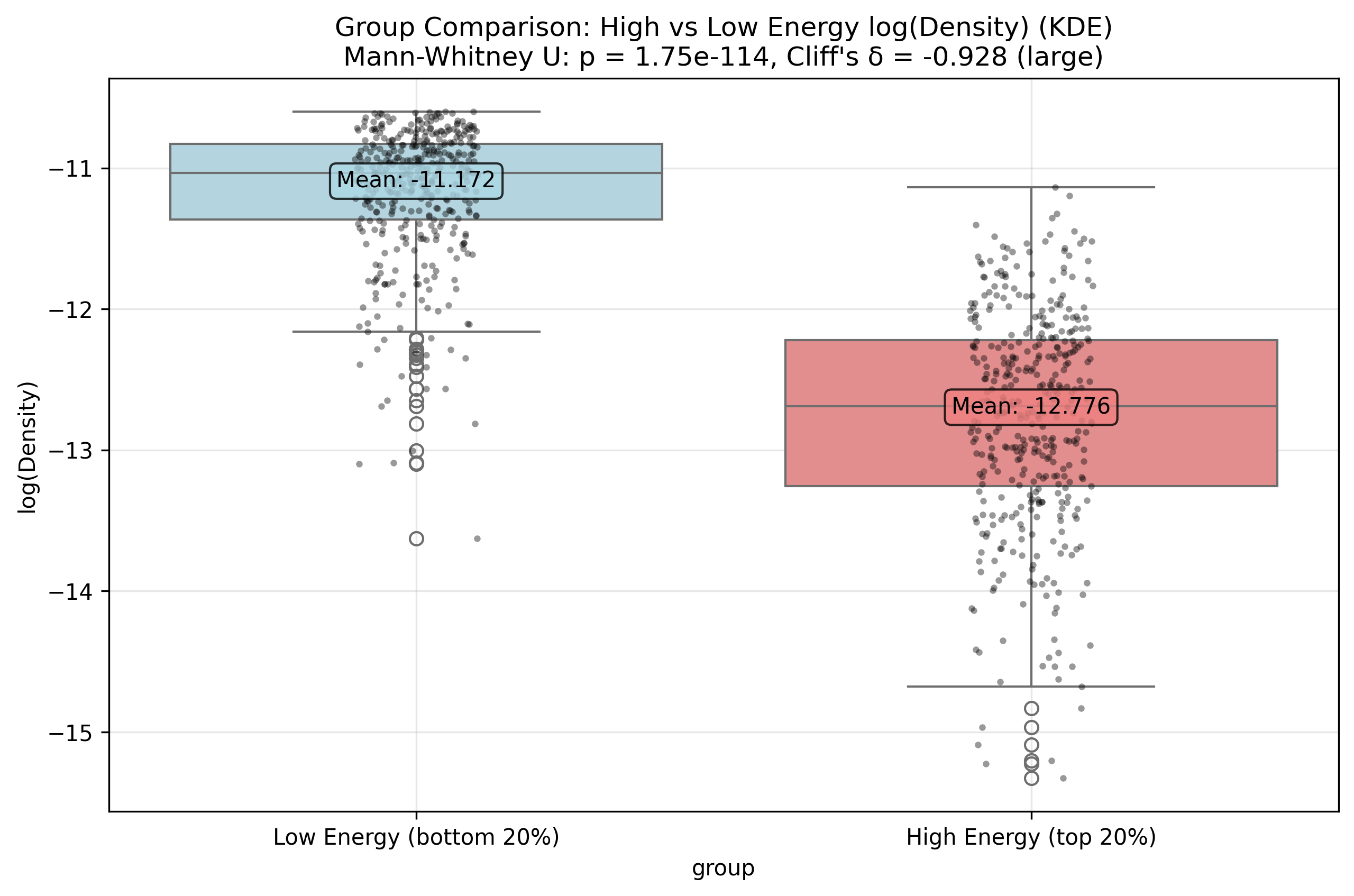}
        \caption{150 steps - KDE ($\delta = -0.928$)}
        \label{fig:group_comparison_150steps_kde}
    \end{subfigure}
    \hfill
    \begin{subfigure}{0.48\textwidth}
        \centering
        \includegraphics[width=\textwidth]{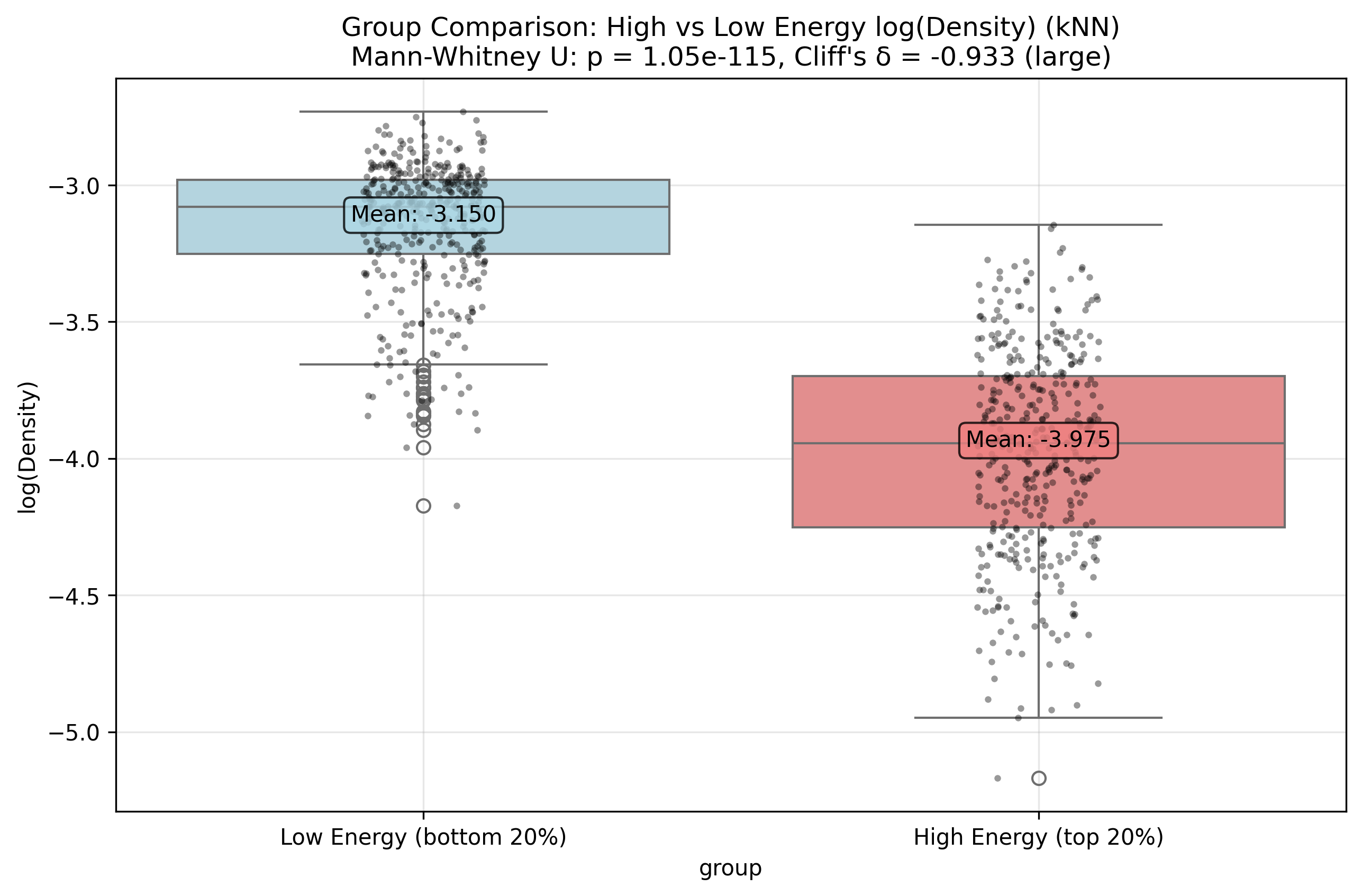}
        \caption{150 steps - k-NN ($\delta = -0.933$)}
        \label{fig:group_comparison_150steps_knn}
    \end{subfigure}

    \caption{Density comparison between high-energy vs. low-energy groups using Cliff's Delta across sampling steps (10, 50, 100, 150 rows) and density estimation methods (KDE left, k-NN right). Large negative effect sizes ($\delta \in [-0.95, -0.87]$) confirm high-energy samples consistently reside in lower-density regions.}
    \label{fig:effect_size_cliffs_delta}
\end{figure}

\begin{figure}[htbp]
    \vspace{-3mm}
    \centering
    \begin{subfigure}{0.4\textwidth}
        \centering
        \includegraphics[width=\textwidth]{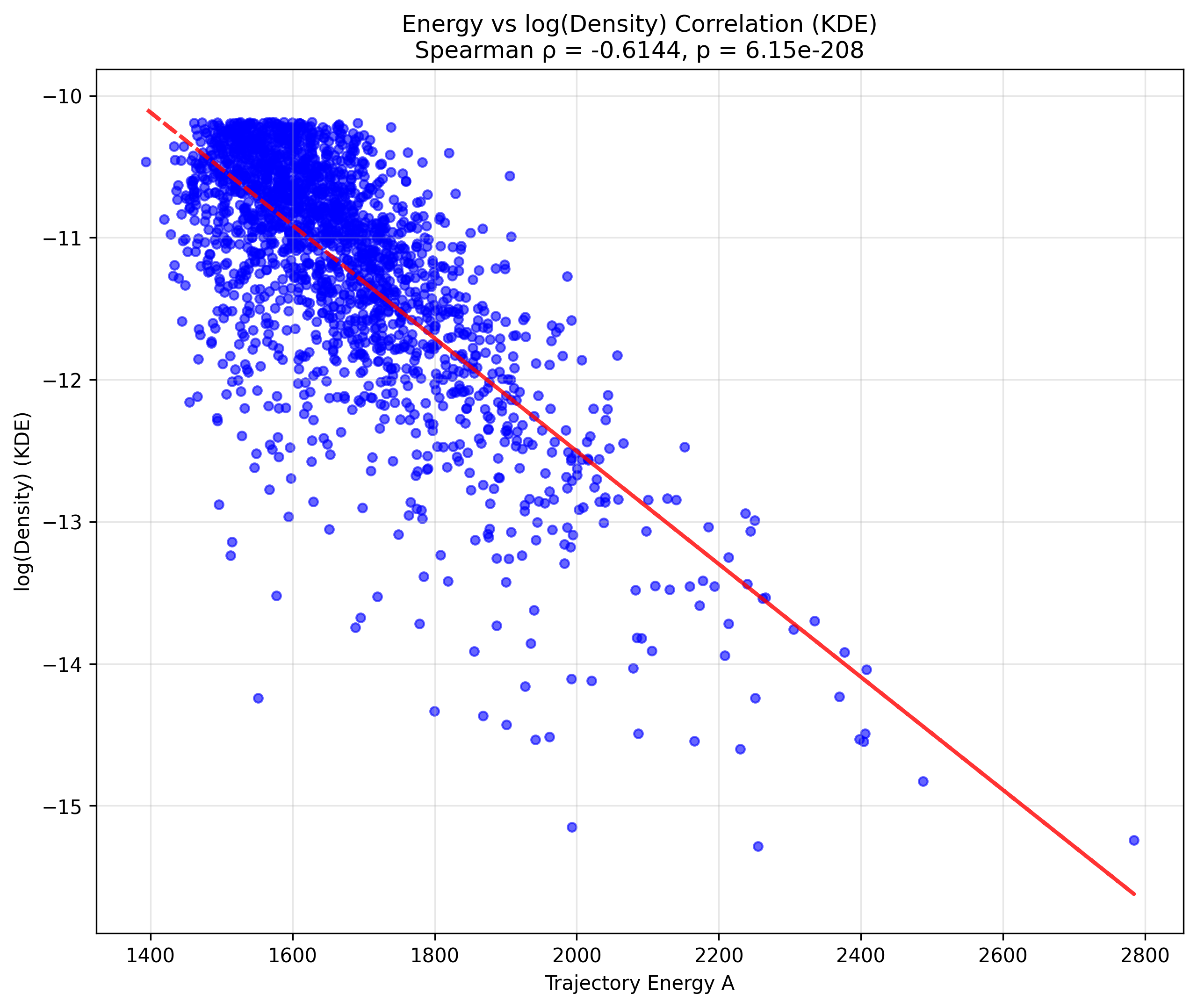}
        \caption{10 steps - KDE ($\rho = -0.6144$, $p = 6.15 \times 10^{-208}$)}
        \label{fig:energy_density_scatter_10steps_kde}
    \end{subfigure}
    \hfill
    \begin{subfigure}{0.4\textwidth}
        \centering
        \includegraphics[width=\textwidth]{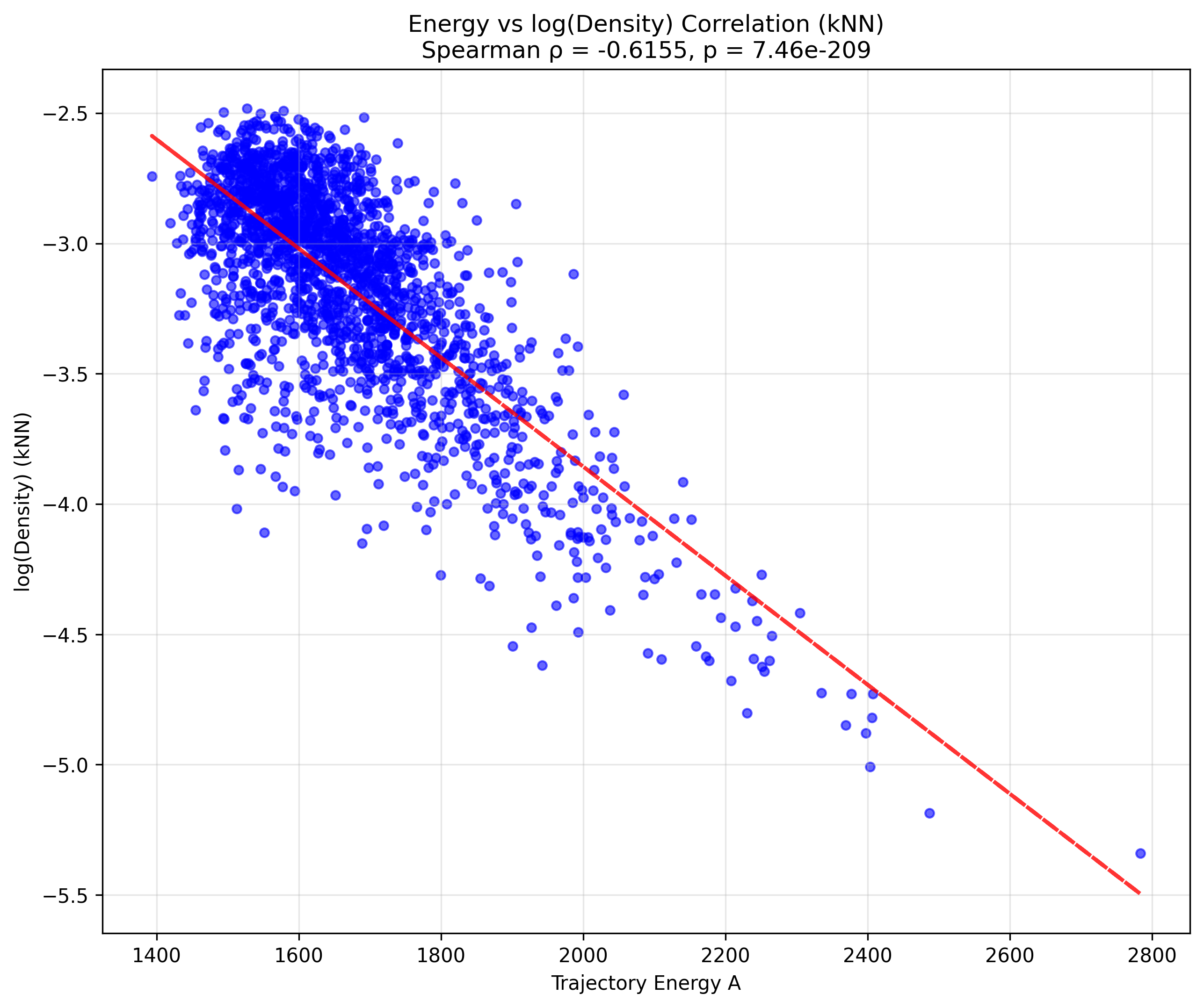}
        \caption{10 steps - k-NN ($\rho = -0.6155$, $p = 7.46 \times 10^{-209}$)}
        \label{fig:energy_density_scatter_10steps_knn}
    \end{subfigure}

    \begin{subfigure}{0.4\textwidth}
        \centering
        \includegraphics[width=\textwidth]{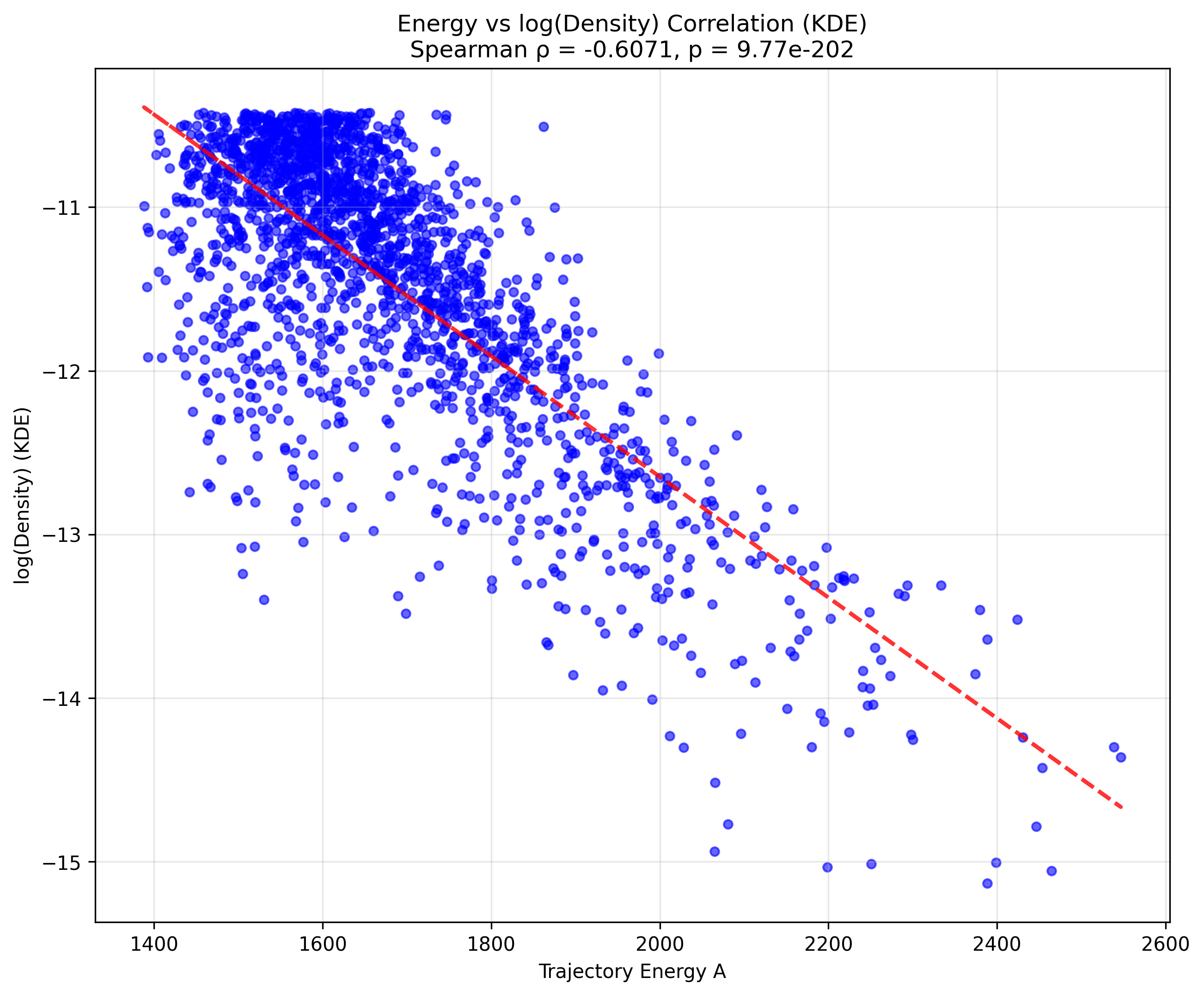}
        \caption{50 steps - KDE ($\rho = -0.6071$, $p = 9.77 \times 10^{-202}$)}
        \label{fig:energy_density_scatter_50steps_kde}
    \end{subfigure}
    \hfill
    \begin{subfigure}{0.4\textwidth}
        \centering
        \includegraphics[width=\textwidth]{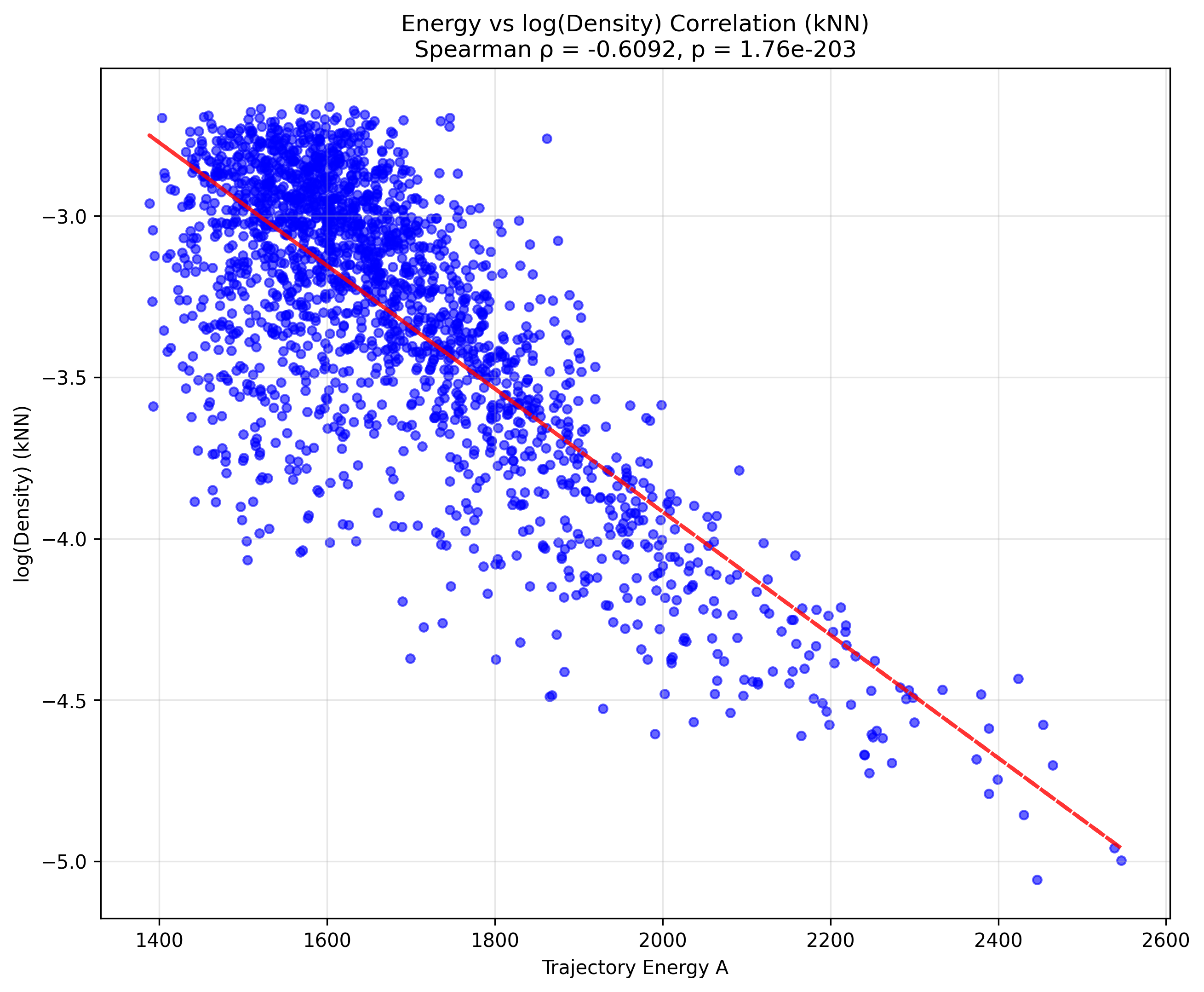}
        \caption{50 steps - k-NN ($\rho = -0.6092$, $p = 1.76 \times 10^{-203}$)}
        \label{fig:energy_density_scatter_50steps_knn}
    \end{subfigure}

    \begin{subfigure}{0.4\textwidth}
        \centering
        \includegraphics[width=\textwidth]{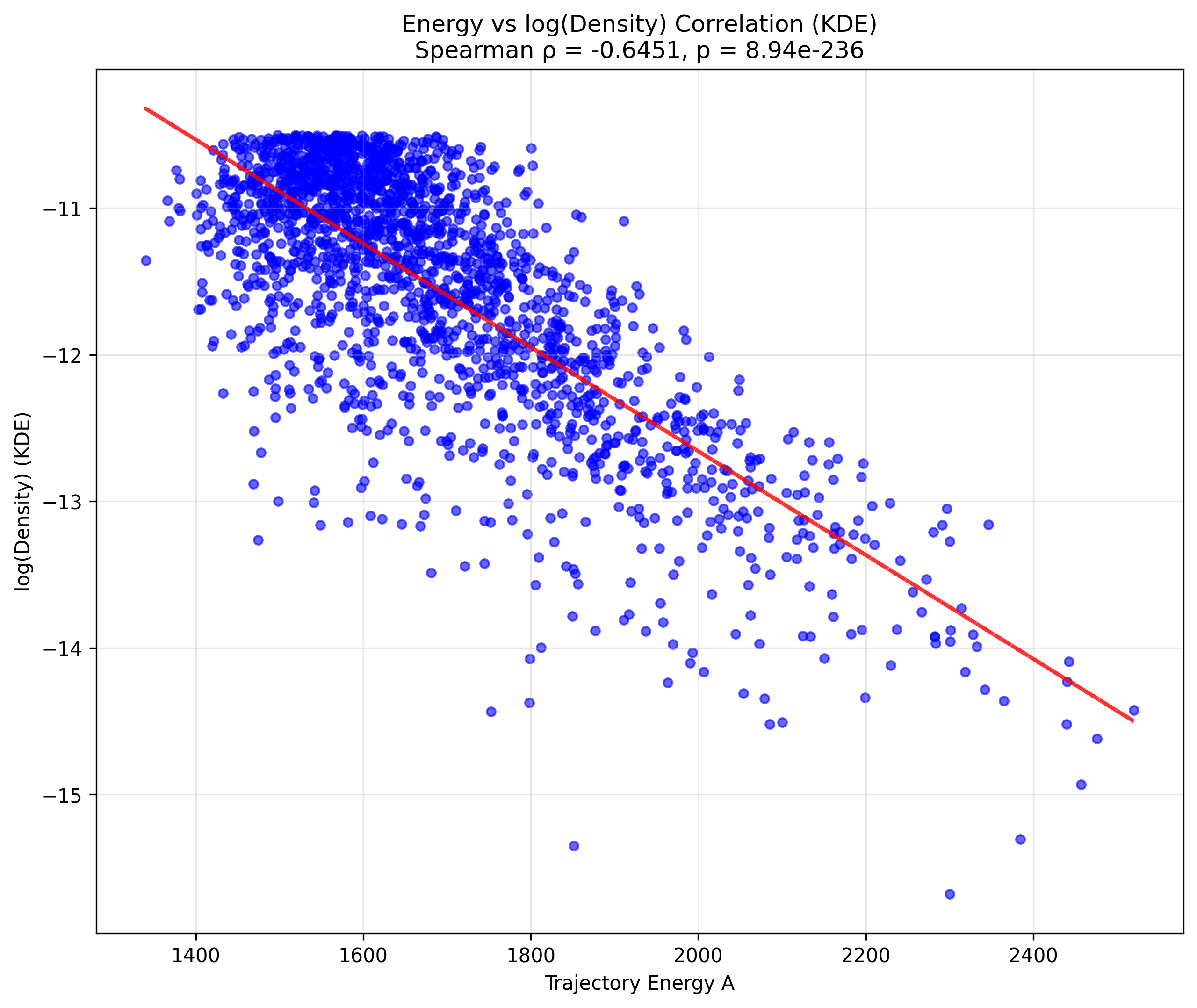}
        \caption{100 steps - KDE ($\rho = -0.6451$, $p = 8.94 \times 10^{-236}$)}
        \label{fig:energy_density_scatter_100steps_kde}
    \end{subfigure}
    \hfill
    \begin{subfigure}{0.4\textwidth}
        \centering
        \includegraphics[width=\textwidth]{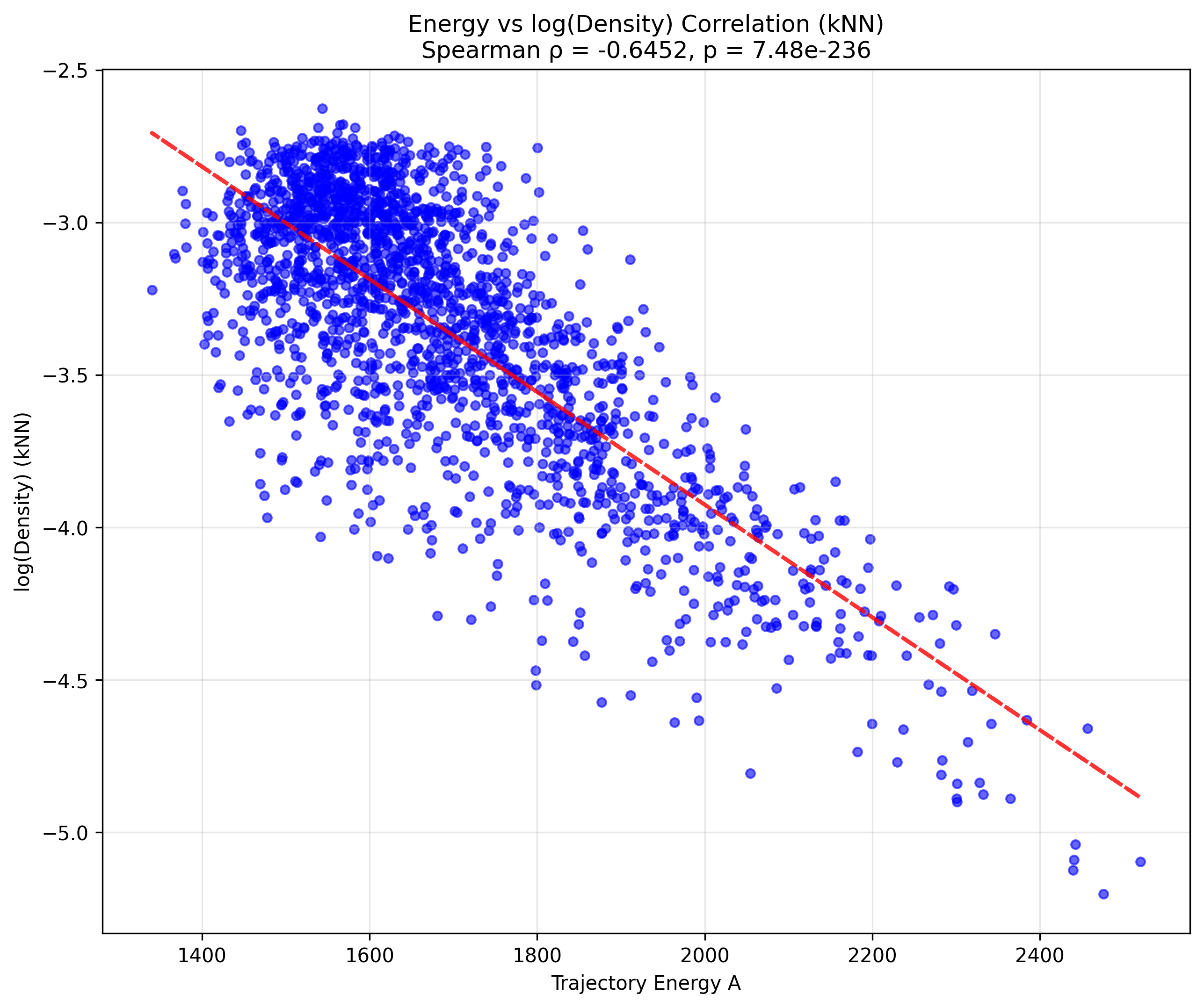}
        \caption{100 steps - k-NN ($\rho = -0.6452$, $p = 7.48 \times 10^{-236}$)}
        \label{fig:energy_density_scatter_100steps_knn}
    \end{subfigure}

    \begin{subfigure}{0.4\textwidth}
        \centering
        \includegraphics[width=\textwidth]{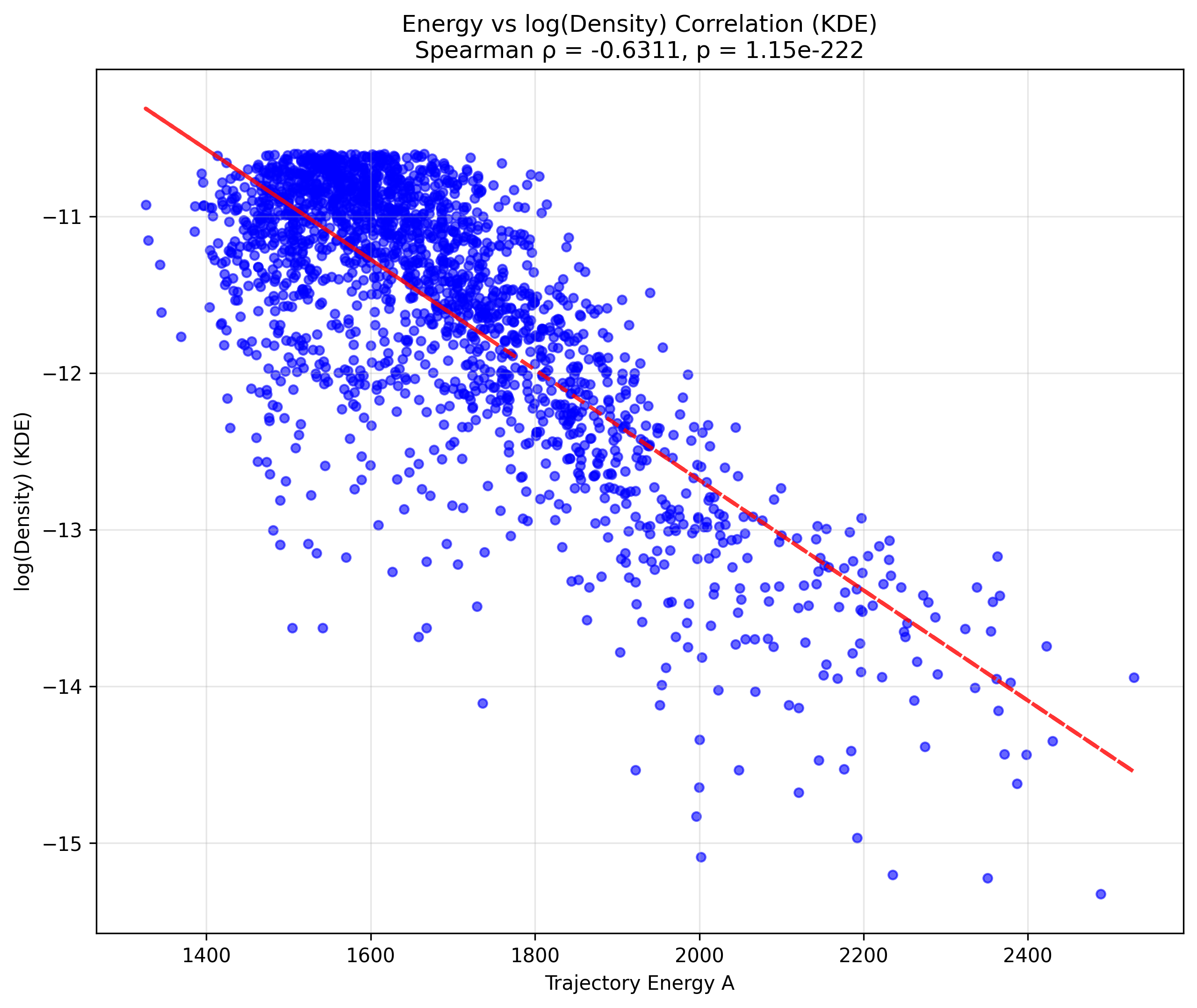}
        \caption{150 steps - KDE ($\rho = -0.6311$, $p = 1.15 \times 10^{-222}$)}
        \label{fig:energy_density_scatter_150steps_kde}
    \end{subfigure}
    \hfill
    \begin{subfigure}{0.4\textwidth}
        \centering
        \includegraphics[width=\textwidth]{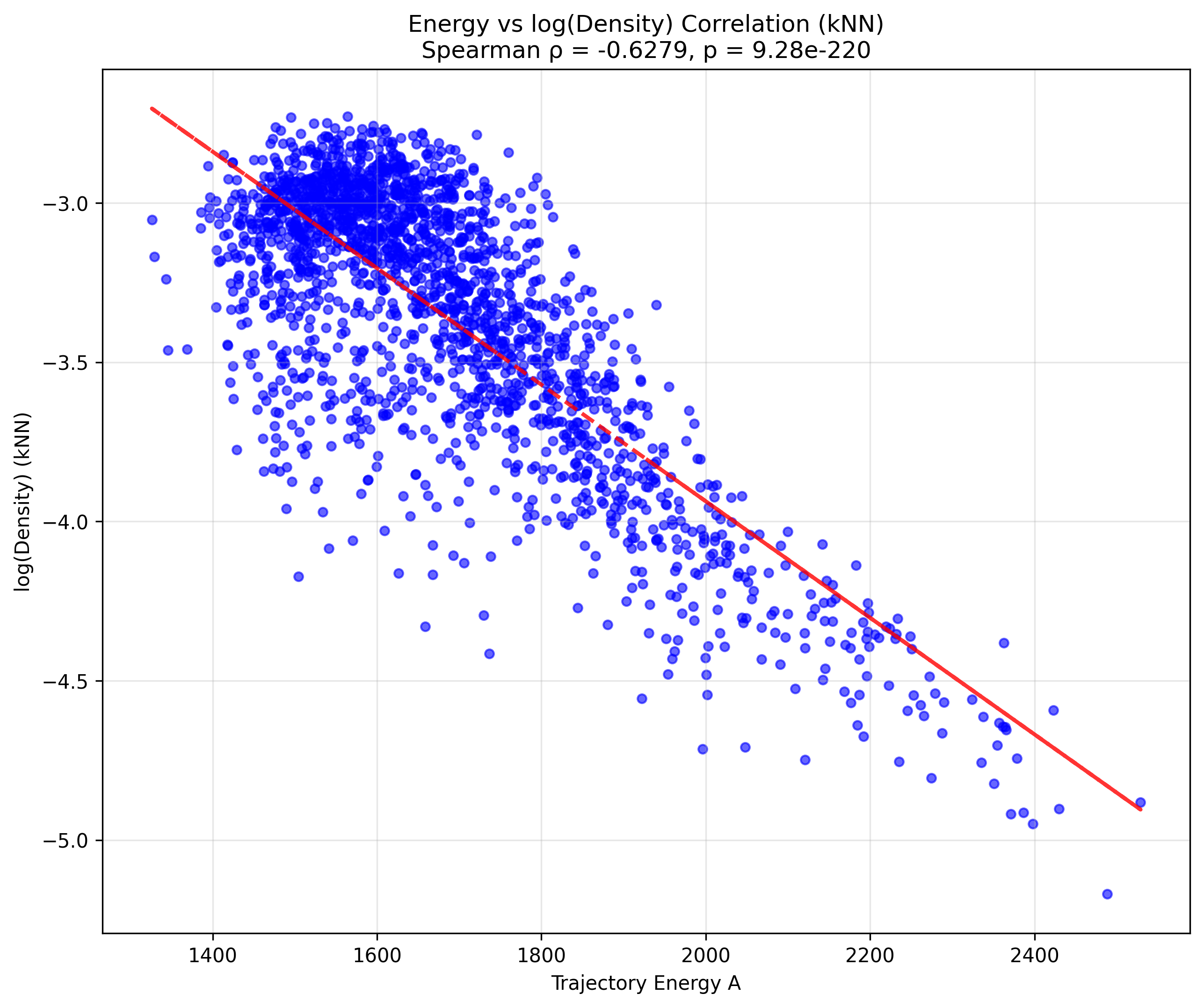}
        \caption{150 steps - k-NN ($\rho = -0.6279$, $p = 9.28 \times 10^{-220}$)}
        \label{fig:energy_density_scatter_150steps_knn}
    \end{subfigure}

    \caption{KPE ${E}$ vs. Data Density correlation across steps (10, 50, 100, 150) using KDE and k-NN. Strong negative correlations (Spearman $\rho \in [-0.65, -0.61]$, $p < 10^{-200}$) confirm higher energy aligns with lower density.}
    \label{fig:energy_density_correlation}
\end{figure}

\begin{figure}[htbp]
    \centering
    \begin{subfigure}{0.48\textwidth}
        \centering
        \includegraphics[width=\textwidth]{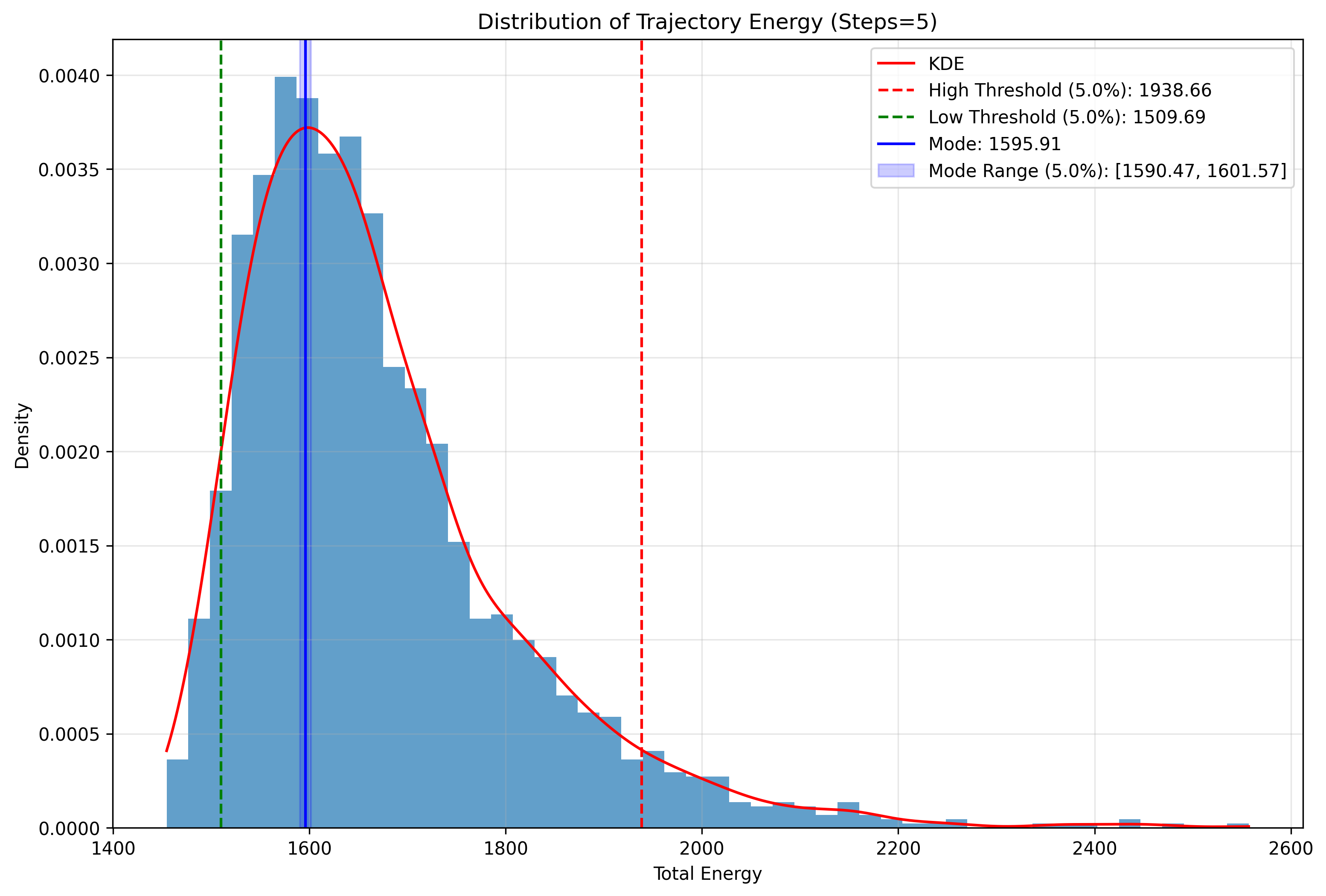}
        \caption{5 sampling steps}
        \label{fig:A_distribution_5steps}
    \end{subfigure}
    \hfill
    \begin{subfigure}{0.48\textwidth}
        \centering
        \includegraphics[width=\textwidth]{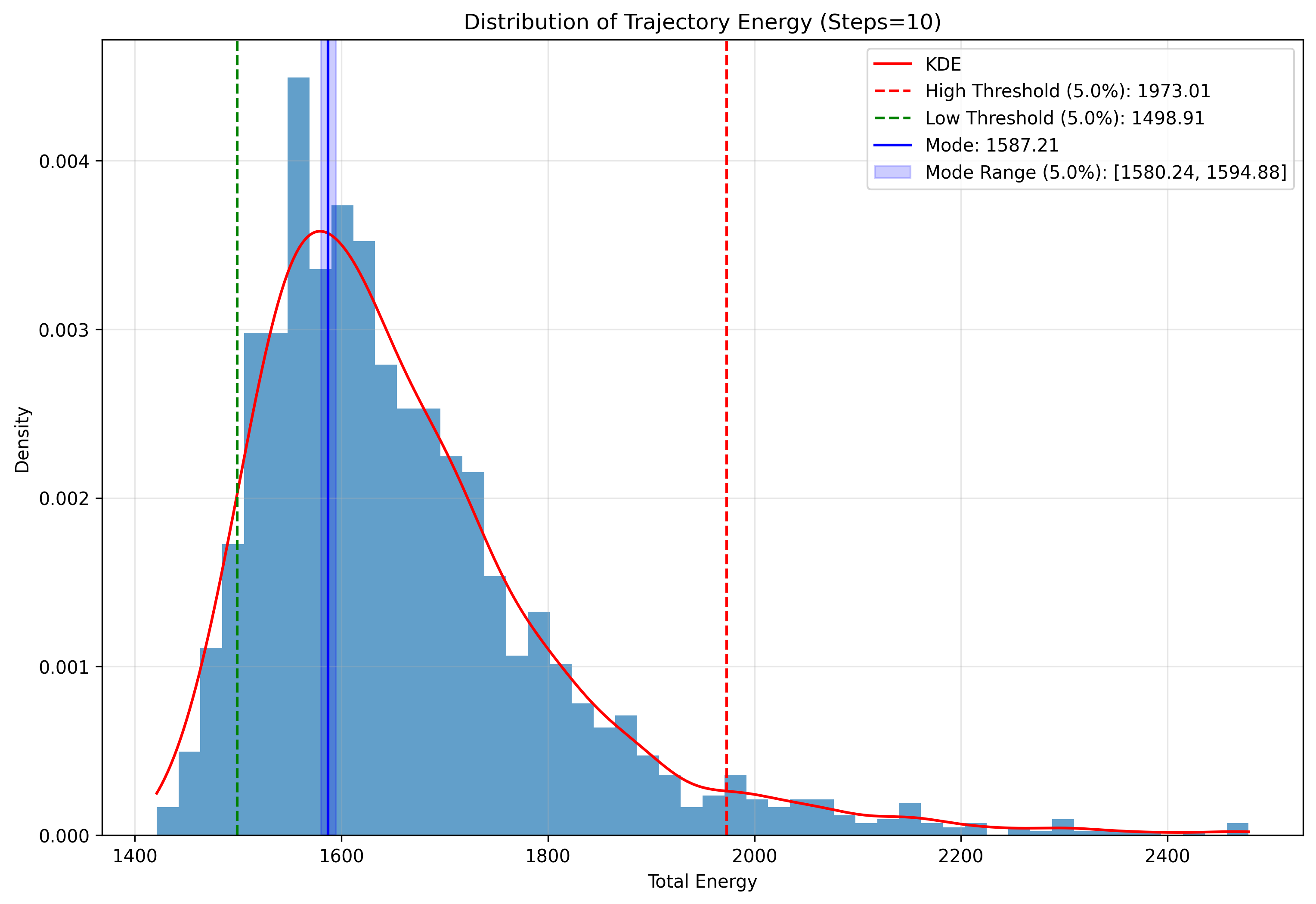}
        \caption{10 sampling steps}
        \label{fig:A_distribution_10steps}
    \end{subfigure}

    \vspace{0.3cm}

    \begin{subfigure}{0.48\textwidth}
        \centering
        \includegraphics[width=\textwidth]{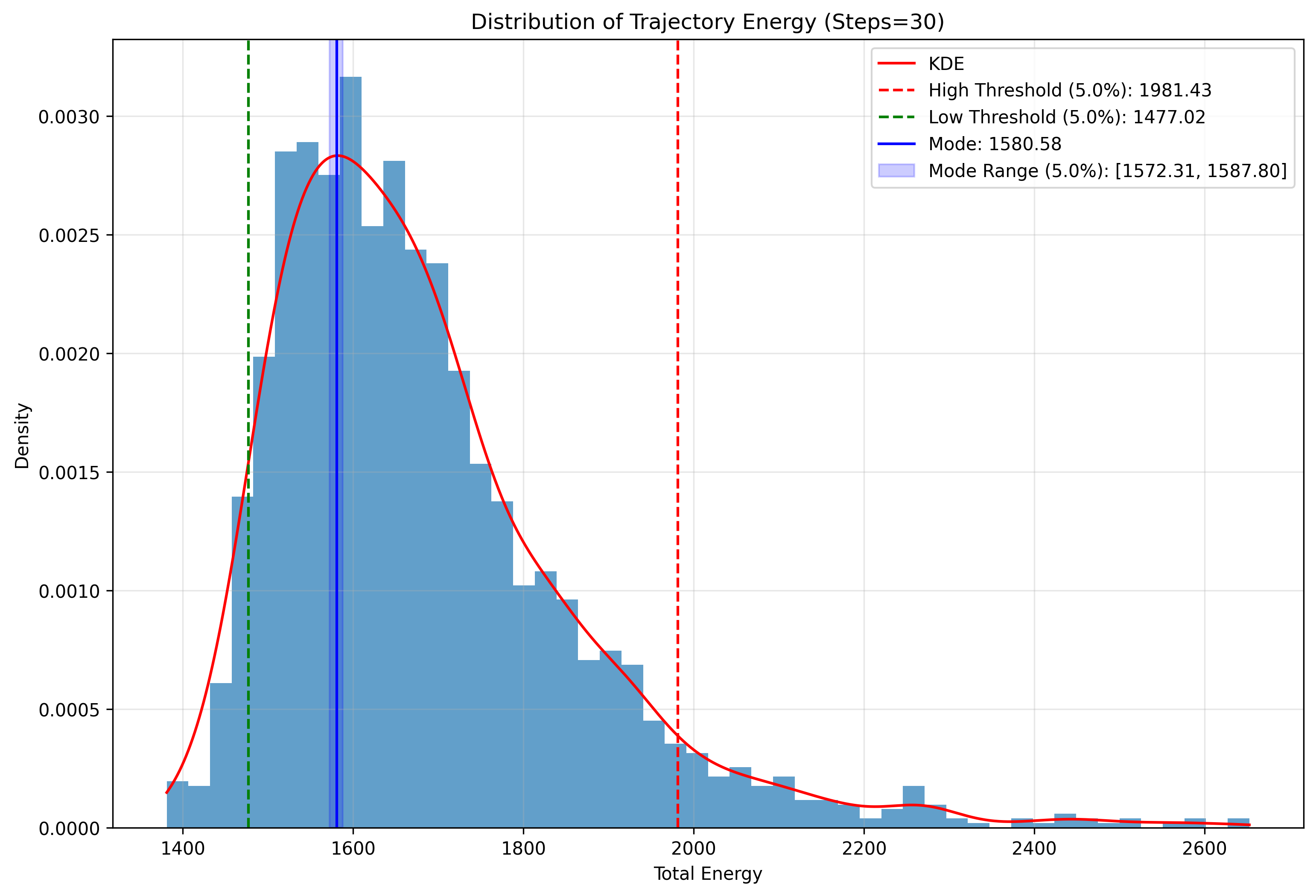}
        \caption{30 sampling steps}
        \label{fig:A_distribution_30steps}
    \end{subfigure}
    \hfill
    \begin{subfigure}{0.48\textwidth}
        \centering
        \includegraphics[width=\textwidth]{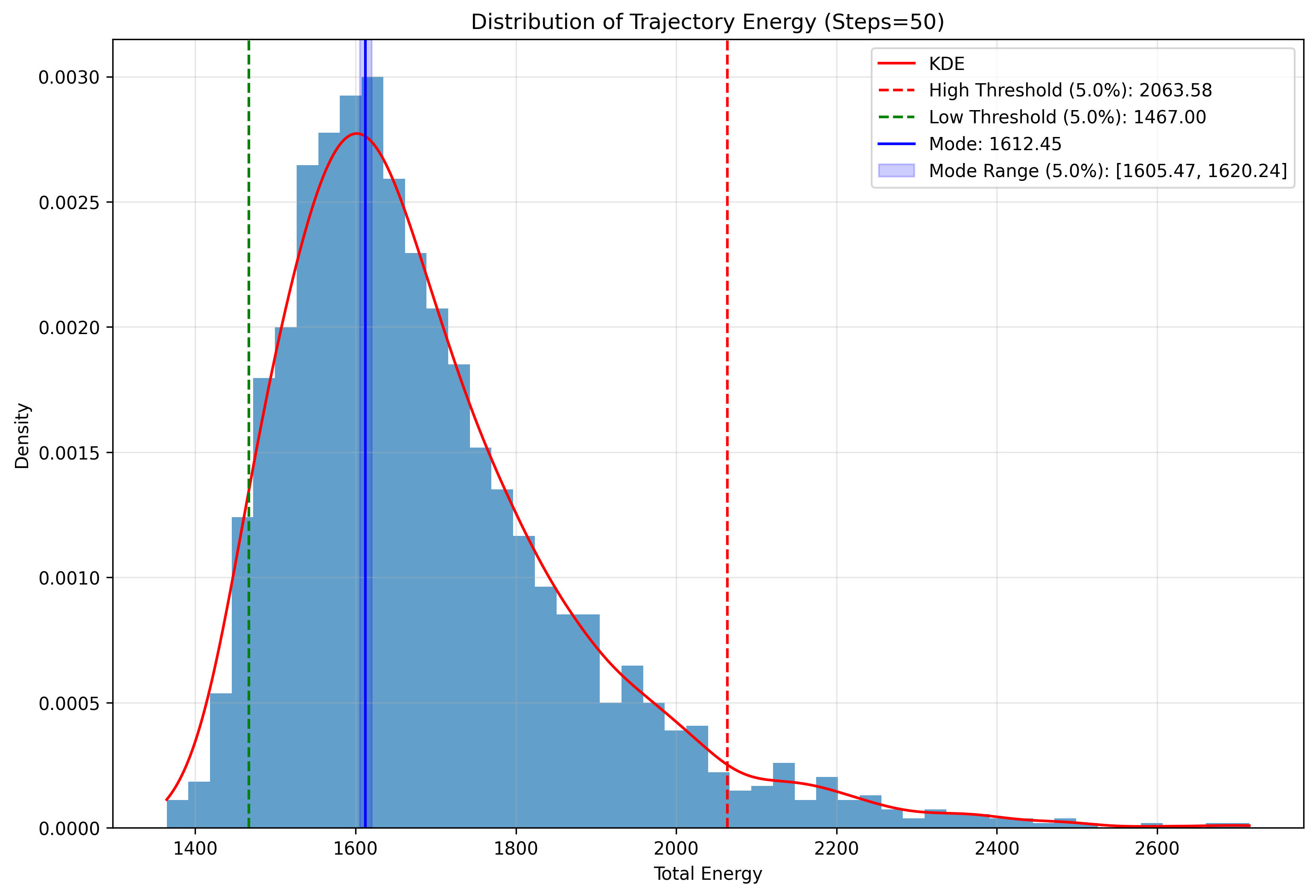}
        \caption{50 sampling steps}
        \label{fig:A_distribution_50steps}
    \end{subfigure}

    \vspace{0.3cm}

    \begin{subfigure}{0.48\textwidth}
        \centering
        \includegraphics[width=\textwidth]{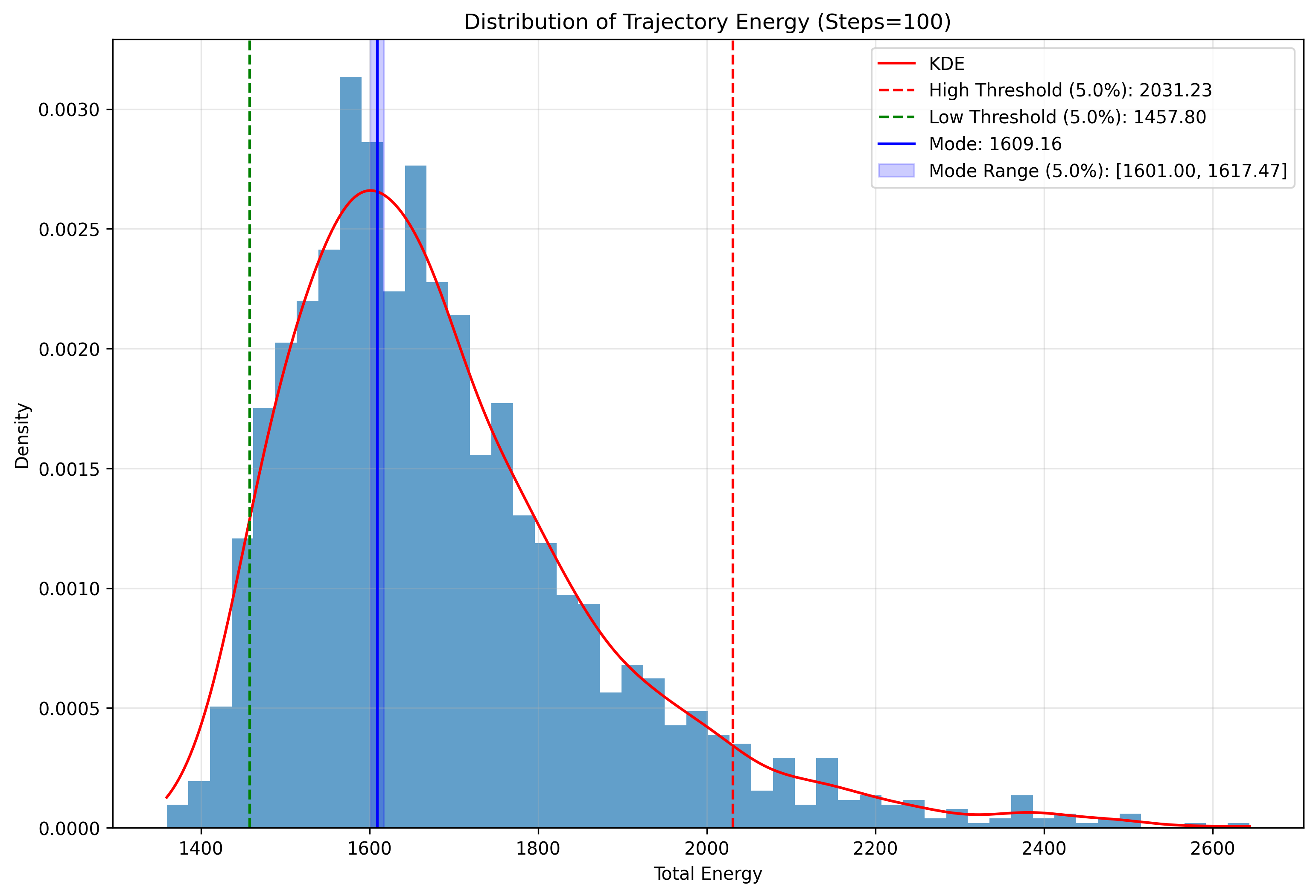}
        \caption{100 sampling steps}
        \label{fig:A_distribution_100steps}
    \end{subfigure}
    \hfill
    \begin{subfigure}{0.48\textwidth}
        \centering
        \includegraphics[width=\textwidth]{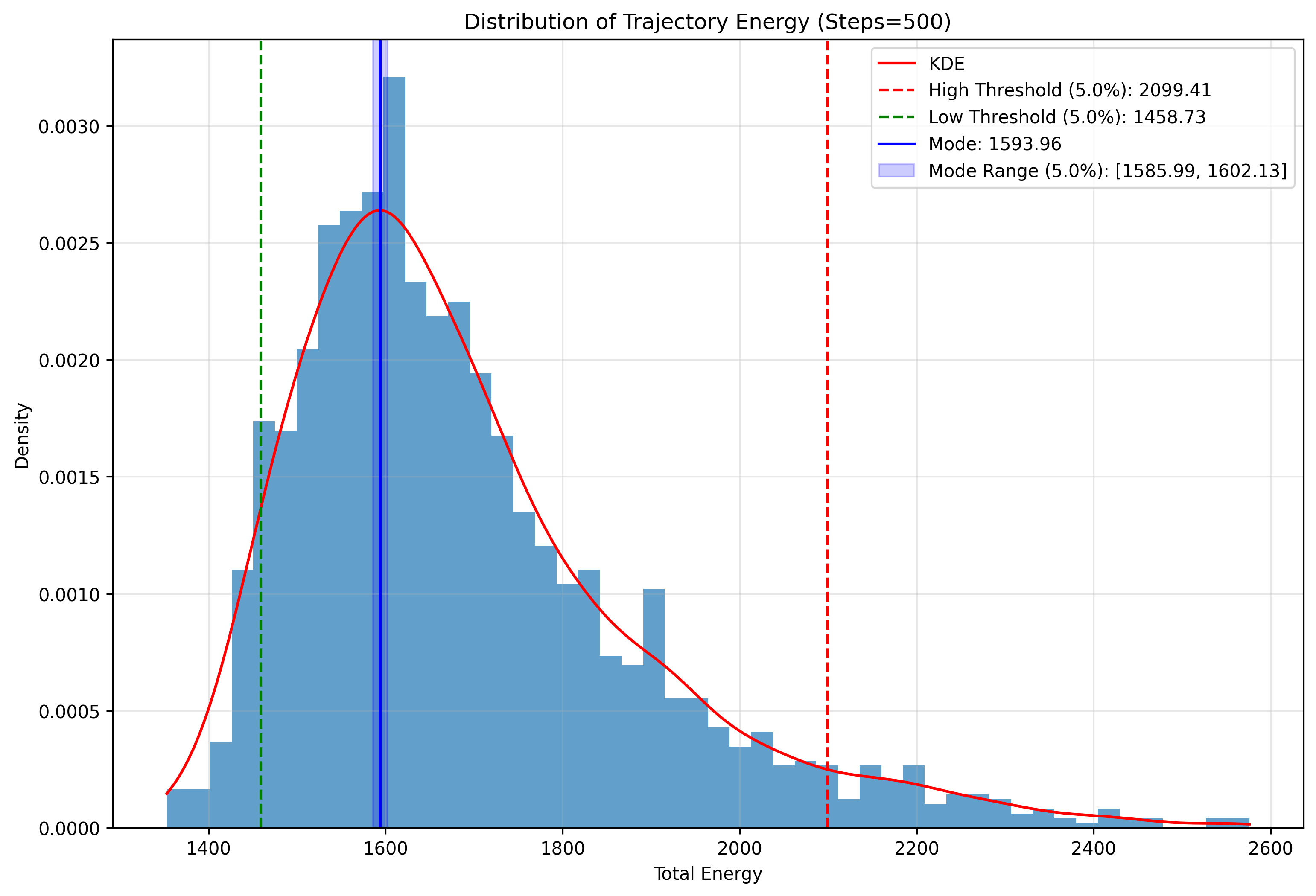}
        \caption{500 sampling steps}
        \label{fig:A_distribution_500steps}
    \end{subfigure}

    \caption{Kinetic path energy (KPE) distributions on CIFAR-10 across varying sampling steps (5, 10, 30, 50, 100, 500). The distributions remain stable with consistent mean ($\mu \approx 1850 \pm 100$) and standard deviation ($\sigma \approx 350 \pm 50$). These observations suggest that KPE is relatively stable across sampling configurations.}
    \label{fig:energy_distribution_robustness}
\end{figure}

\end{document}